\documentclass[10pt,twocolumn,letterpaper]{article} 

\usepackage[pagenumbers]{cvpr} 
\usepackage{times}
\usepackage{epsfig}
\usepackage{graphicx}
\usepackage{amsmath}
\usepackage{amssymb}
\usepackage{xcolor}         
\usepackage{float}
\usepackage{caption}
\usepackage{amsfonts,amssymb}
\usepackage{mathrsfs}
\usepackage{tabularx}
\usepackage{wrapfig}
\usepackage{verbatim}
\usepackage{dsfont}
\usepackage{tablefootnote}
\usepackage[stable]{footmisc}
\usepackage{booktabs}
\usepackage{multicol}
\usepackage{multirow}
\usepackage{color}
\usepackage{subcaption}
\usepackage{pythonhighlight}
\usepackage{algpseudocode}
\usepackage{algorithm}
\usepackage{physics}

\newcommand{\rulesep}{\color{black} \unskip\ \vrule\ }

\definecolor{todocolor}{RGB}{255,0,00}
\newcommand\TODO[1] {\PackageWarning{}{Unprocessed todo}\emph{\textcolor{todocolor}{TODO: #1}}}

\definecolor{jiapeng}{rgb}{0.2, 0.4,0.9}

\newcommand{\ANGIE}[1]{{\textbf{\textcolor{blue}{Angie: #1}}}}

\newcommand{\set}[1]{\mathcal{#1}}
\newcommand{\expnumber}[2]{{#1}\mathrm{e}{#2}}
\DeclareMathOperator{\MLP}{MLP}

\DeclareMathOperator{\NC}{NC}
\DeclareMathOperator{\RC}{RC}
\DeclareMathOperator{\glo}{glo}
\DeclareMathOperator{\anc}{anc}
\DeclareMathOperator{\sym}{sym}
\DeclareMathOperator{\usym}{usym}
\DeclareMathOperator{\flip}{flip}

\renewcommand{\paragraph}[1]{\smallskip\noindent\textbf{#1}}

\definecolor{cvprblue}{rgb}{0.21,0.49,0.74}
\usepackage[pagebackref,breaklinks,colorlinks,citecolor=cvprblue]{hyperref}

\def\paperID{3729} 
\def\confName{CVPR}
\def\confYear{2024}

\title{DPHMs: Diffusion Parametric Head Models for Depth-based Tracking }

\author{
    Jiapeng Tang$^1$ \quad Angela Dai$^1$ \quad Yinyu Nie$^1$ \quad Lev Markhasin$^2$ \quad Justus Thies$^3$ \quad Matthias Nie{\ss}ner$^1$ \\
    \\
     $^{1}$ Technical University of Munich \quad
     $^{2}$ Sony Semiconductor Solutions Europe \\
     $^{3}$  Technical University of Darmstadt  \\
      \url{https://tangjiapeng.github.io/projects/DPHMs} \\
}

\begin{document}
\twocolumn[{%
\renewcommand\twocolumn[1][]{#1}%
\maketitle
\begin{center}
    \vspace{-0.55cm}
    \centering
    \captionsetup{type=figure}
    \includegraphics[width=\linewidth]{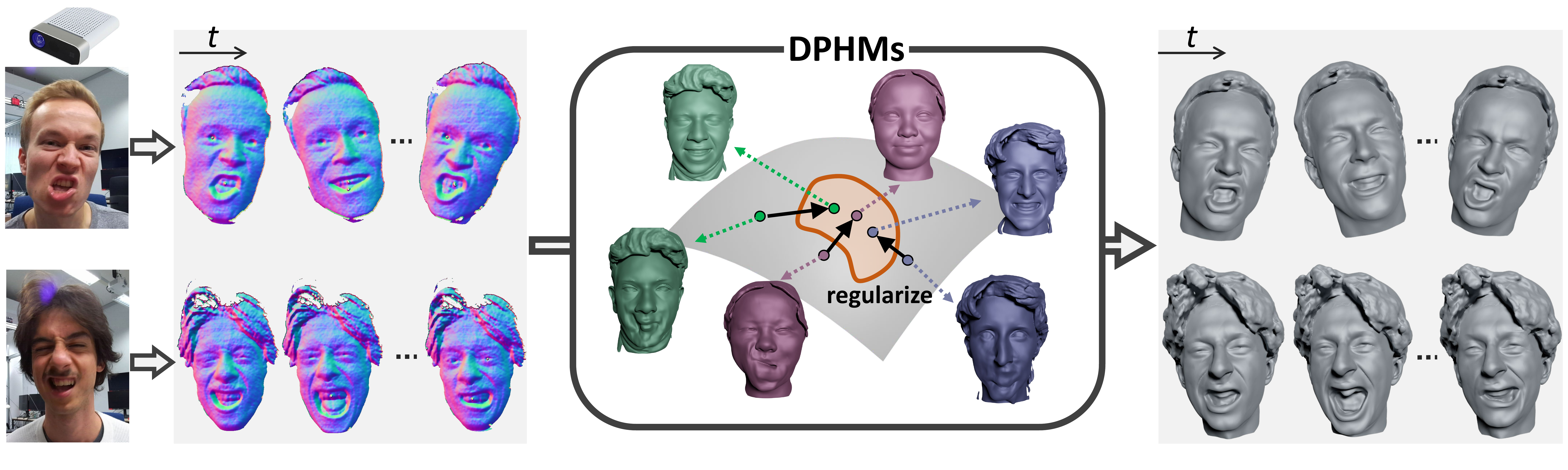}
    \caption{
    We present DPHMs, a diffusion parametric head model that is used for robust head reconstruction and expression tracking from monocular depth sequences. 
    Leveraging the DPHM diffusion prior, we effectively constrain the identity and expression codes on the underlying latent manifold when fitting to noisy and partial observations of commodity depth sensors.
    }
    \vspace{0.25cm}
    \label{fig:teaser}
\end{center}
}]
\begin{abstract}
%
We introduce Diffusion Parametric Head Models (DPHMs), a generative model that enables robust volumetric head reconstruction and tracking from monocular depth sequences. 
While recent volumetric head models, such as NPHMs, can now excel in representing high-fidelity head geometries, tracking and reconstructing heads from real-world single-view depth sequences remains very challenging, as the fitting to partial and noisy observations is underconstrained. 
To tackle these challenges, we propose a latent diffusion-based prior to regularize volumetric head reconstruction and tracking.
%
%
This prior-based regularizer effectively constrains the identity and expression codes to lie on the underlying latent manifold which represents plausible head shapes.
To evaluate the effectiveness of the diffusion-based prior, we collect a dataset of monocular Kinect sequences consisting of various complex facial expression motions and rapid transitions.
We compare our method to state-of-the-art tracking methods and demonstrate improved head identity reconstruction as well as robust expression tracking. 
%
\end{abstract}
\section{Introduction}
\label{sec:intro}
The fascination with 3D models of human heads spans across millennia, evolving from sculptural artistry to the realm of computer graphics.
Creating a digital twin for each person is poised to revolutionize entertainment and communication, potentially transforming applications in video calls, augmented reality, virtual reality, animated movies, and gaming. 
For this revolution, we have to use affordable hardware that is accessible to everyone like webcams or Kinect-like depth sensors that are built into smartphones and laptops.
Especially to capture the metrical size of a human, the depth sensors play an important role, as actual distances are measured.
However, with a single depth sensor, only a part of the person is visible and the visible part contains sensor and surface-dependent noise.
This leads to a generally under-constraint reconstruction problem which needs to be addressed with data priors~\cite{blanz2023morphable,paysan20093d,cao2013facewarehouse, booth20163d, booth20173d, ploumpis2019combining,bolkart2015groupwise, brunton2014multilinear,li2017learning}.
The most common data priors for face reconstruction are so-called 3D morphable models (3DMM)~\cite{blanz2023morphable,li2017learning} which capture facial shape and expression variations using Principal Component Analysis.
However, these 3DMMs use a fixed template mesh which restricts their capacity to model the full landscape of head identity geometries and expressions (\eg, diverse hairstyles, wrinkles, intricate faces, etc.). 
Recently, Neural Parametric Head Models (NPHMs) \cite{giebenhain2023learning} have overcome these limitations by modeling full-head avatars with a broad spectrum of hair geometries and intricate non-linear facial deformations through an over-parameterized coordinate MLP-based neural field. 
This over-parametrization, however, is a key limitation when NPHMs are used for an underconstrained reconstruction task.
Noisy or sparse input data leads to severe overfitting of NPHMs, with highly unrealistic head reconstructions.
%

To this end, we introduce DPHMs, the first diffusion generative model designed to generate clean and diverse 3D heads from noisy NPHM latent representations. 
Our key idea is to couple NPHMs with denoising diffusion models that can produce high-fidelity and diverse samples by navigating in the latent space.
We learn identity and expression parametric diffusion through iterative transitions between noisy and clean latent representations, over-parametrized by NPHMs. 
We leverage the noise estimation during diffusion of the DPHM model to represent the gradient of the identity and expression latent distributions, enabling effective regularization of the identity and expression while fitting to real sequences for commodity depth sensors.
As illustrated in Fig.~\ref{fig:teaser}, when fitting NPHMs to noisy and partial data, latent vectors might fall outside the underlying latent surface manifold, generating implausible head geometries.
Using our DPHM prior, those latents can be regularized towards latent vectors that are on the surface manifold, generating high-quality head geometries.
%
%

We evaluate our proposed method on a new challenging benchmark that contains various extreme facial expression motions with rapid transitions captured with a monocular Kinect Azure sensor.
Extensive experiments and comparisons against recent state-of-the-art head reconstruction methods demonstrate that our approach can reconstruct more accurate head geometries and expressions and achieve more robust and coherent facial expression tracking.
%

\medskip
\noindent
Our contributions can be summarized as follows: 
\begin{itemize}
    \item We propose the first diffusion generative model that creates clean and diverse 3D heads by explicitly learning the distributions of identity and expression latent defined in neural parametric head models.
    \item We design novel regularization terms based on diffusion parametric head models, effectively constraining the latent optimization when fitting sparse and noisy observations from monocular depth sequences.
    \item We collect a dataset of monocular Kinect scan sequences with various challenging facial expression motions for evaluation benchmark.
\end{itemize}
\section{Related Work}
\label{sec:related}

\paragraph{3D morphable face and head models.}
 The conception of 3D morphable face and head models can be dated back to the pioneering work of Blanz and Vetter~\cite{blanz2023morphable}. They introduced the concept of a 3D Morphable Face Model (3DMM) based on a dataset of 200 3D face scans and used Principal Component Analysis (PCA) to represent facial shape and texture variations compactly.
 To enhance expressiveness, Cootes et al. introduced the Active Appearance Model (AAM)\cite{cootes1998active}. Some subsequent models incorporated more captured data~\cite{paysan20093d,cao2013facewarehouse, booth20163d, booth20173d, ploumpis2019combining}.
 Advanced facial models were designed to go beyond linear spaces. These include multi-linear models~\cite{bolkart2015groupwise, brunton2014multilinear}, non-linear models~\cite{tran2018nonlinear}, and the FLAME model~\cite{li2017learning}, which seamlessly integrates linear shape spaces with articulated head components.
Recently, researchers have explored integrating Signed Distance Functions (SDFs) and deformation fields for human faces~\cite{giebenhain2023learning, giebenhain2024mononphm, zheng2022imface, yenamandra2021i3dmm}, bodies~\cite{tang2021learning, palafox2021npms, palafox2022spams}, and animals~\cite{lei2022cadex, tang2022neural}. While these neural parametric models excel in producing high-fidelity geometries and estimating complex non-linear deformations, they often struggle to generate reasonable samples from random noise. In contrast, our approach enhances neural parametric models with diffusion models, effectively mapping random noise latent vectors onto the desired surface manifold.

\paragraph{Head reconstruction and tracking.}
Building upon the data priors of 3D morphable models, many works~\cite{thies2016face2face, tuan2017regressing, tewari2017mofa, deng2019accurate, RingNet:CVPR:2019, guo2020towards, shang2020self, Feng:SIGGRAPH:2021, EMOCA:CVPR:2021, Zielonka2022TowardsMR, grassal2022neural} tried to reconstruct 3D faces from monocular images or videos. 
Our work is more closely related to those endeavors focusing on the 3D face/head reconstruction from scans~\cite{lin2005markerbased,zhang2005markerless, bradley2010markerless}. For a comprehensive overview of early face tracking from scan sequences, we recommend referring to \cite{pighin2006trackingreview}.
An early attempt at tracking using commodity depth sensors can be found in \cite{weise2011kinectearly}, where pre-recorded animation priors were applied. Li~\cite{li2013kinect} proposed a real-time method for Kinect tracking. Thies et al. integrated RGBD face tracking with facial reenactment in \cite{thies2015reenactment}.
More recently, several works~\cite{chan2021pigan, chan2022eg3d, deng2022gram, xiang2023gramhd, an2023panohead, ramon2021h3dnet, lin2023ssif, zheng2022imavatar, zheng2023pointavatar} have attempted to reconstruct human heads using coordinate-MLP representations.
A concurrent work of MonoNPHM~\cite{giebenhain2024mononphm} extends NPHMs for head tracking from monocular RGB videos. 
IMAvatar \cite{zheng2022imavatar} and PointAvatar~\cite{zheng2023pointavatar} utilize coordinate-MLP to build personalized canonical head geometries, which are tracked through skinning weight fields generalized from the FLAME model.
However, these methods struggle to reconstruct high-quality head avatars with intricate details due to a lack of effective geometry priors. In contrast, we learn high-quality priors from high-resolution full-head scans.
\paragraph{Diffusion models.}
Denoising Diffusion Models (DDMs)~\cite{sohl2015deep,song2019generative,song2020improved,song2020denoising,ho2020denoising} have shown unprecedented generation diversity and realism in various data domains, including images~\cite{ho2020denoising, meng2021sdedit, kim2022diffusionclip, nichol2021glide, avrahami2022blended, saharia2022image, ho2022cascaded,  dhariwal2021diffusion, rombach2022high,lugmayr2022repaint}, videos~\cite{he2022lvdm,text2video-zero,xing2023make,chen2023videodreamer}, and shapes~\cite{luo2021diffusion,zhou20213d,zeng2022lion,zhang20233dshape2vecset,hui2022neural,tang2024diffuscene,cao2024motion2vecsets,zhang2023functional}
To synthesize high-dimensional data effectively, some methods compress physical data into a high-dimensional latent space~\cite{rombach2022high, stablediffusion, zeng2022lion, zhang20233dshape2vecset,tang2024diffuscene,cao2024motion2vecsets} and learn the distribution of latent features. Our approach adopts a similar strategy, parametrizing high-resolution head avatar geometry into two separate latent spaces (identity and expression) and leveraging latent diffusion models to learn their data distributions.
Some recent works leverage the learned priors from diffusion models to guide the NeRF optimization~\cite{mildenhall2021nerf}, such as DreamFusion~\cite{poole2022dreamfusion}, DiffusionNeRF~\cite{wynn2023diffusionerf}, and Single Stage NeRF~\cite{chen2023ssdnerf}. 
%
A concurrent work of FaceTalk~\cite{aneja2023facetalk} uses diffusion models to synthesize temporally consistent 3D motion sequences of human heads based on NPHMs~\cite{giebenhain2023learning}.
In this work, we design diffusion prior regularizers to effectively constrain the identity and expression codes to lie on the underlying latent manifold that represents plausible
head shapes, producing robust head reconstruction tracking when fitting to noisy and partial scans of commodity depth sensors.
\section{Approach}
\label{SecApp}
\begin{figure}
    \vspace{-3mm}
    \centering
    \includegraphics[width=.95\linewidth]{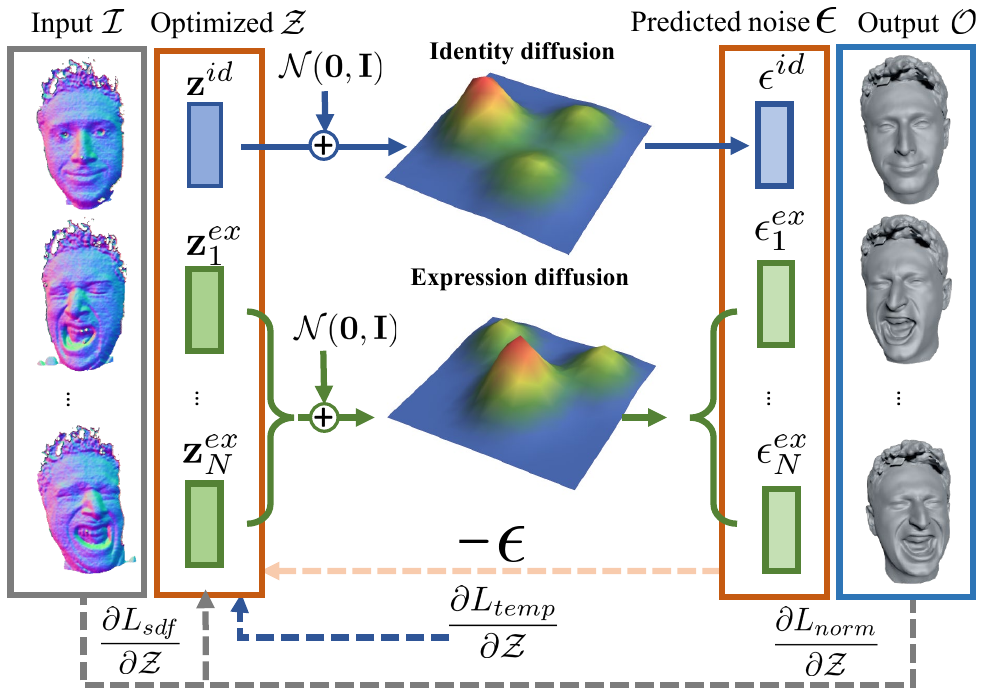}
    \caption{\textbf{DPHMs for depth-based tracking.} 
    Given a sequence of depth maps $\mathcal{I}$ of N frames, our objective is to reconstruct a full-head avatar $\mathcal{O}$ including its expression transitions. 
    To achieve this, we optimize the parametric latent $\mathcal{Z} = \{ \Vec{z}^{id}, \Vec{z}^{ex}_1, ..., \Vec{z}^{ex}_N \}$ of NPHM that can be decoded into continuous signed distance fields $\mathcal{O}$ by identity and expression decoders.
    To align with the observations, we calculate data terms $L_{sdf}$ and $L_{norm}$ between $\mathcal{I}$ and $\mathcal{O}$. However, high-level noise still makes navigating the latent optimization extremely challenging.
    At the core of our method is an effective latent regularization using diffusion priors; we add Gaussian noises to $\mathcal{Z}$ and then pass them into identity and expression diffusion models to predict perturbed noise $\mathcal{\epsilon}$ for updating $\mathcal{Z}$.
    The diffusion regularizer guides $\Vec{z}^{id}$ and ${\Vec{z}^{ex}_i}$ towards the individual manifold  of their distributions via $\epsilon^{id}$ and $\epsilon^{ex}$, ensuring plausible head geometry reconstruction and robust tracking.
    To enhance temporal coherence, $L_{temp}$ penalizes inconsistency between ${\Vec{z}^{ex}_i}$ of nearby frames.
    }
    \label{fig:overview}
    \vspace{-3mm}
\end{figure}
Given a monocular sequence of depth maps $\mathcal{I}$, we aim to reconstruct a series of full-head avatars $\mathcal{O}$, see Fig.~\ref{fig:overview}.
To accurately reconstruct various facial expressions with topological variations, we choose to predict a continuous signed distance field for each frame using a modified NPHM model, which is parametrized by $\Vec{z}^{id}$ and $\Vec{z}^{ex}_i$.
However, the latent optimization to real-world single-view depth sequences is extremely challenging, as the fitting to partial and noisy observations is underconstrained.
To overcome these challenges, we introduce Diffusion Parametric Head Models (DPHMs), the first diffusion generative model tailored for generating clean and diverse 3D heads from noisy latent representations. 
It is used as a prior to regularize the NPHM identity and expression codes which significantly improves head tracking robustness from noisy and partial scans.
%
%
%
%
%
\subsection{Diffusion Parametric Head Model}
\label{SubSecAppDPHM}
Our Diffusion Parametric Head Model (DPHM) is designed to learn a diffusion generative model to enable robust head tracking and reconstruction.
We build a disentangled latent space of shape and expression, inspired by Neural Parametric Head Models (NPHMs)~\cite{giebenhain2023learning}.
NPHMs represent head geometries using an SDF decoder in canonical space and capture facial expressions through forward deformations. However, this approach maintains the same mesh connectivity in the canonical identities, limiting the ability to change topologies during expression tracking. To address this limitation, we replace forward deformations with backward deformations, which learn the deformation fields from arbitrary expressions to canonical space. This enables the reconstruction of a continuous signed distance field for each expression space by wrapping points from the expression to canonical space and querying SDF values.
Please refer to the supplementary material for more details about the revised NPHM model based on backward deformations.
%
%
\subsubsection{NPHM Identity Space}
\label{SubSubSecAppIden}
Following NPHMs~\cite{giebenhain2023learning}, we represent the signed distance field of canonical identity by an ensemble of several smaller
local MLP-based networks individually responsible for local regions centered at 39 pre-defined anchors of human heads. Concretely, we define $K = 2 K_{\sym} + K_{\usym}$ facial anchors, denoted as $\Vec{a} \in \mathbb{R}^{K \times 3}$, which are estimated by a small $\MLP_{\anc}$ from the global latent $\Vec{z}^{id}_{\glo}$.
$K_{\sym}$ anchors are on the left face. They are mirrored to the other $K_{\sym}$ ones on the right face.
$K_{\usym}$ anchors are in the middle of the face, shared by both the left and right faces. For $k_{th}$ local region of facial anchor, we represent its local geometry via a local latent vector $\Vec{z^{id}_{k}}$, along with the global latent vector $\Vec{z^{id}_{\glo}}$ as well as an SDF decoder $\MLP_{\Vec{\theta}_k}$ parametrized by learnable weights $\Vec{\theta}_k$:
\vspace{-2mm}
\begin{equation}
    f_k (\Vec{p}, \Vec{z^{id}_k}, \Vec{z^{id}_{glo}}) = \MLP_{\theta_k} ([\Vec{p} - \Vec{a}_k, \Vec{z^{id}_k}, \Vec{z^{glo}} ]) .
\end{equation}
Finally, we can composite local fields into a global field:
\vspace{-2mm}
\begin{equation}
    F_{id} (\Vec{p}) = \sum_{k=1}^{K} \omega_k(\Vec{p}, \Vec{a_k}) f_k (\Vec{p}, \Vec{z^{id}_k}, \Vec{z^{id}_{glo}})  .
\end{equation}
The blending weights are calculated by a Gaussian kernel based on the Euclidean distance between the query point $\Vec{p}$ and $\Vec{a_k}$.
%
\subsubsection{Backward-Deformation Expression Space}
\label{SubSubSecAppDeform}
In contrast to NPHMs~\cite{giebenhain2023learning}, which define their expression space through forward deformations, limiting topology to that of the canonical shape, we model the expression space through globally conditioned backward deformation fields learned by a $\MLP_\psi$~\cite{giebenhain2023learning, aneja2023facetalk}.
We use a latent expression vector $\Vec{z}^{ex}$ to explain the geometry variations caused by expression transitions. Since such a
deformation is also closely related to the identity geometry, 
the deformation decoder also receives the identity code $\Vec{z}^{id} = \Vec{z}^{id}_{glo} \oplus \Vec{z}^{id}_{1} \cdots \oplus \Vec{z}^{id}_{K} $ as an additional condition. 
 To enforce the deformation network to learn a disentangled expression representation independent of the identity latent, we impose a constraint that neutral expressions for the canonical space are close to zeros:
 \vspace{-2mm}
\begin{equation}
    F_{ex} (\Vec{p}) = \MLP_{\phi}([\Vec{p}, \Vec{z}^{id}, \Vec{z}^{ex} ]).
    \vspace{-1mm}
\end{equation}
We jointly train the identity and expression networks in an auto-decoder fashion. Once finished, we can obtain a set of identity latents $\mathcal{Z}^{id} = \{ \Vec{z}^{id}_j \}_j$ for canonical geometries and a set of expression latents $\mathcal{Z}^{ex} = \{ \Vec{z}^{ex}_{j,l} \}_{j,l}$, where $\Vec{z}^{ex}_{j,l}$ means the $l_{th}$  expression of $j_{th}$ subject.
\subsubsection{Neural Parametric Diffusion}
\label{SubSecAppParacDiff}
In this section, we will explain how DPHMs are learned and their connection with score functions.   
To learn the identity and expression latent diffusion, we consider them as flattened 1D latent vectors. 
The identity and expression diffusion models are treated analogously; the only difference is the number of input and output channels in the denoising network. In the following, we will use the expression latent diffusion to explain the details.
Given an expression code $\Vec{x}_0$ sampled from $\mathcal{Z}^{ex}$ over-parametrized from the training dataset.  
The forward diffusion process progressively adds Gaussian noise to $\Vec{x}_0$, obtaining a series of corrupted versions $\Vec{x}_1, ..., \Vec{x}_T$, according to a linearly increased noise variance schedule $\beta_1, ..., \beta_T$  along the step $t$ ( $\beta_1 < ... < \beta_T$).
The diffusion step at time $t$ is defined as:
\vspace{-2mm}
\begin{equation}
    \label{Equadiffusion_each}
    \Vec{x}_{t} := \sqrt{ 1-\beta_{t} } \Vec{x}_{t-1}   +  \beta_{t} \Vec{\epsilon}_{t-1} ,
    \vspace{-1mm}
\end{equation}
where $\epsilon_{t-1} \sim \mathcal{N}(\Vec{0},\Vec{I})$.
%
%
With the re-parametrization trick, we obtain:
\vspace{-2mm}
\begin{equation}
    \label{Equadiffusion_repara}
    \Vec{x}_t := \sqrt{\bar{\alpha}_t} \Vec{x}_0 + \sqrt{1-\bar{\alpha}_t} \Vec{\epsilon} ,
    \vspace{-1mm}
\end{equation}
where $\epsilon \sim \mathcal{N}(\Vec{0},\Vec{I})$, $\alpha_t := 1 - \beta_t$  , $\bar{\alpha}_t := \prod_{r=1}^{t} \alpha_s$.
The reverse diffusion process is tasked to remove the noise gradually. In each denoising step $t$, we have:
\vspace{-2mm}
\begin{equation}
    \label{Equadenoise_each}
    p_\Vec{\phi} (\Vec{x}_{t-1} | \Vec{x}_{t} ) :=  \mathcal{N}(\Vec{x}_{t-1}; \Vec{\mu}_{\Vec{\phi}}(\Vec{x}_{t}, t), \Tilde{\beta_t}(\Vec{I}) ) ,
    \vspace{-1mm}
\end{equation}
where $\Tilde{\beta_t} := (1 - \overline{\alpha}_{t-1}) \beta_t / (1-\overline{\alpha}_t) $.
Instead of directly predicting $\Vec{\mu}_{\Vec{\phi}}(\Vec{x}_{t}, t)$, we opt to predict the noise $\Vec{\epsilon}_{\phi}(\Vec{x}_{t}, t)$ using the denoiser $\Vec{\epsilon}_{\phi}$ and the noise data $\Vec{x}_{t}$.
Then, $\Vec{\mu}_{\Vec{\phi}}(\Vec{x}_{t})$ can be re-parametrized by subtracting the predicted noise:
%
\vspace{-2mm}
\begin{equation}
    \label{Equareparam}
    \Vec{\mu}_{\Vec\phi}(\Vec{x}_{t}, t) :=  \frac{1}{\sqrt{\alpha_{t}}} (\Vec{x}_{t} - \frac{\beta_t}{\sqrt{1-\bar{\alpha}_t}} \Vec{\epsilon}_{\Vec{\phi}}(\Vec{x}_{t}, t)) .
    \vspace{-1mm}
\end{equation}
By training the denoiser on various noise levels and time steps, we obtain the empirical risk expectation as the loss:
\vspace{-2mm}
\begin{equation}
    \label{equaNoisypred}
    \begin{aligned}
    \mathbb{E}_{\Vec{x}_0, \Vec{\epsilon}, t} [ \frac{\beta_t}{ 2\alpha_t (1-\overline\alpha_t) }  \| \Vec{\epsilon} - \Vec{\epsilon}_{\Vec{\phi}} ( \sqrt{\bar{\alpha}_t} \Vec{x}_0 +  \sqrt{1-\bar{\alpha}_t} \Vec{\epsilon}, t) \|^2] .
    \end{aligned}
    \vspace{-1mm}
\end{equation}
As depicted in~\cite{vincent2011connection}, a DDM noise
estimator has a connection to score matching~\cite{hyvarinen2005estimation, song2019generative, song2020denoising, song2020improved, song2020score, song2020sliced, song2021maximum} and is proportional to the score function:
\vspace{-2mm}
\begin{equation}
    \label{equaScore}
    \Vec{\epsilon}_{\Vec{\phi}} (\Vec{x}_t, t)  \propto -\nabla_{\Vec{x}} \log p(\Vec{x}).
    \vspace{-1mm}
\end{equation}
Therefore, moving in the opposite direction of the noise predicted by the neural network is akin to approaching the modes of the data distribution.
This principle can be harnessed for generating samples that closely resemble the data distribution through the Langevin dynamics~\cite{welling2011bayesian,sohl2015deep}.
\subsection{DPHM for Monocular Depth-based Tracking} 
\label{SubSecAppTracking}
Given a depth map sequence $\mathcal{I} = \{ [\Vec{P}_i,  \Vec{N}_i ]\}_{i=1}^{N}$ of $N$ frames, where $\Vec{P}_i,  \Vec{N}_i$ are back-projected points and normals of $i_{th}$ frame, 
our goal is to discover the identity vector $\Vec{z}^{id}$ along with N expressive latent vectors $\Vec{z}^{ex}_{1:N}$ that can be interpreted into full-head avatars $\mathcal{O}$ with hairs and accurate expression transitions through the pre-trained identity and deformation decoders in Sec.~\ref{SubSubSecAppIden} and ~\ref{SubSubSecAppDeform}.
According to Bayes's theorem, it is equivalent to maximizing the $\emph{posterior}$ probability of identity and expression latent $\set{Z} = \{ \Vec{z}^{id}, \Vec{z}^{ex}_{1:N} \}$. 
This can be further formulated as: 
\vspace{-2mm}
\begin{equation}
    \label{equaRecPosterior}
    \begin{aligned}
         %
        p(\set{Z}  | \set{I} )  =  \frac{p(\set{I} | \set{Z} )  p( \set{Z} )} { p(\set{I}) }
    \end{aligned}
    \vspace{-1mm}
\end{equation}
In practice, we can maximize the log-posterior as follows:
\vspace{-2mm}
\begin{equation}
    \label{equaRecLogPosterior}
    \begin{aligned}
        %
        \log p(\set{Z} | \set{I} )  =  \log p(\set{I} | \set{Z}  ) + \log p( \set{Z} ) - \log p(\set{I})  \\
        \ge \log p(\set{I} | \set{Z}  ) + \log p( \Vec{z}^{id}) +  \log (\Vec{z}^{ex}_{1:N}) .
    \end{aligned}
    \vspace{-1mm}
\end{equation}
As $\set{I}$ is neither independent on $\Vec{z}^{id}$ nor $\Vec{z}^{ex}_{1:N}$, we can drop $\log p(\set{I})$ that is a normalizing constant. Based on the assumption that identity and expression are disjoint, we can consider $\Vec{z}^{id}$ and $\Vec{z}^{ex}_{1:N}$ are independent. We can update $\set{Z}$ with stochastic gradient descent:
\vspace{-2mm}
\begin{equation}
    \label{equaRecLogPosterior}
    \begin{aligned}
        \nabla_{\set{Z}} \log p(\set{Z} | \set{I} )  = 
        \nabla_{\set{Z} } \log p( \set{I} | \set{Z} ) + \nabla_{\Vec{z}^{id}} \log p( \Vec{z}^{id}) \\ + \nabla_{\Vec{z}^{ex}_{1:N}} \log p( \Vec{z}^{ex}_{1:N}) ,
    \end{aligned}
    \vspace{-1mm}
\end{equation}
with the first term being the gradient of the log-likehood $\log p( \set{I} | \set{Z} ) $, the other two terms are the gradient of the log-prior $\log p( \Vec{z}^{id}) $ and $\log p( \Vec{z}^{ex}_{1:N})$ respectively.
$\nabla_{\Vec{z}^{id}} \log p( \Vec{z}^{id})$ is exactly the score function of identity diffusion of DPHMs, while the log-prior of all expression latents within a sequence $\nabla_{\Vec{z}^{ex}_{1:N}} \log p( \Vec{z}^{ex}_{1:N}) \propto \sum_{i=1}^{N} \log p( \Vec{z}^{ex}_i)$ which is the sum of score function of our expression latents for all observed frames.

By plugging Eq.~\ref{equaScore} into ~\ref{equaRecLogPosterior}, we can use the learned prior of our DPHMs over $\Vec{z}^{id}, \Vec{z}^{ex}_{i}$, and the temporal smoothness constrain between $\Vec{z}^{ex}_{i}$ and $\Vec{z}^{ex}_{i-1}$ to approximate the depencies of near-by frames.
For the first term of $\log p( \set{I} | \set{Z} )$, we use the data terms by considering SDF prediction and normal in-consistency errors of observed points from $\set{I}$.
Thus, the gradient with respect to $\set{Z}$ while minimizing the following loss function $L$ is obtained by:
\vspace{-2mm}
\begin{equation}
    \label{equaRecLoss}
    \begin{aligned}
      \nabla L =  \nabla L_{sdf}(\set{I}) + \lambda_{norm} \nabla L_{norm}(\set{I}) - \lambda_{id} \epsilon^{id}( \hat{\Vec{z}}^{id}) \\
       - \lambda_{ex}  \sum_{i=1}^{N} \epsilon^{ex}(\hat{\Vec{z}}^{ex}_i) + \lambda_{temp} \nabla L_{temp}(\Vec{z}^{ex}_{1:N}).
    \end{aligned}
    \vspace{-1mm}
\end{equation}
note that $\hat{\Vec{z}}$ is the corrupted version of $\Vec{z}$ by adding Gaussian noise at diffusion step $t$ defined in Eq.~\ref{Equadiffusion_repara}.
$L_{sdf}(\set{I})$ enforces a constraint that requires observed points from the $i_{th}$ frame $\Vec{P}_i$, when transformed back into the canonical space to get $\Vec{P}^{id}_i$, to be precisely positioned on the zero iso-surface of the identity geometry:
\vspace{-2mm}
\begin{equation}
    \label{equaRecSDFLoss}
    \begin{aligned}
      L_{sdf}(\set{I}) = \sum_{i=1}^{N} | F_{id}(\Vec{P}^{id}_i, \Vec{z}^{id}) - 0| \\
      \Vec{P}^{id}_i =  F_{ex}(\Vec{P}_i, \Vec{z}^{id}, \Vec{z}^{ex}_i). \quad \quad \quad
    \end{aligned}
    \vspace{-1mm}
\end{equation}
To further attain the facial geometry details, the normal inconsistency is penalized:
\vspace{-2mm}
\begin{equation}
    \label{equaRecNormLoss}
    \begin{aligned}
    L_{norm}(\set{I}) = \sum_{i=1}^{N} < \Vec{N}^{ex}_i, \Vec{N}_i > ,
    \end{aligned}
    \vspace{-1mm}
\end{equation}
where $\Vec{N}^{ex}_i$ is the gradient of the SDF values $F_{id}(\Vec{P}^{id}_i, \Vec{z}^{id}) $ with respect to $\Vec{P}_i$, based on the theorem in IGR~\cite{gropp2020implicit}.
The temporal smoothness term for the expressions is defined as:
\vspace{-2mm}
\begin{equation}
    \label{equaRecTempSmoLoss}
    \begin{aligned}
      L_{temp}(\Vec{z}^{ex}_{1:N}) = \sum_{i=2}^{N} \|\Vec{z}^{ex}_{i} - \Vec{z}^{ex}_{i-1}\|_2^2 .
    \end{aligned}
    \vspace{-1mm}
\end{equation}

\section{DPHM-Kinect Dataset}
\begin{figure}[h]
\vspace{-3mm}
    \centering
    \includegraphics[width=\linewidth]{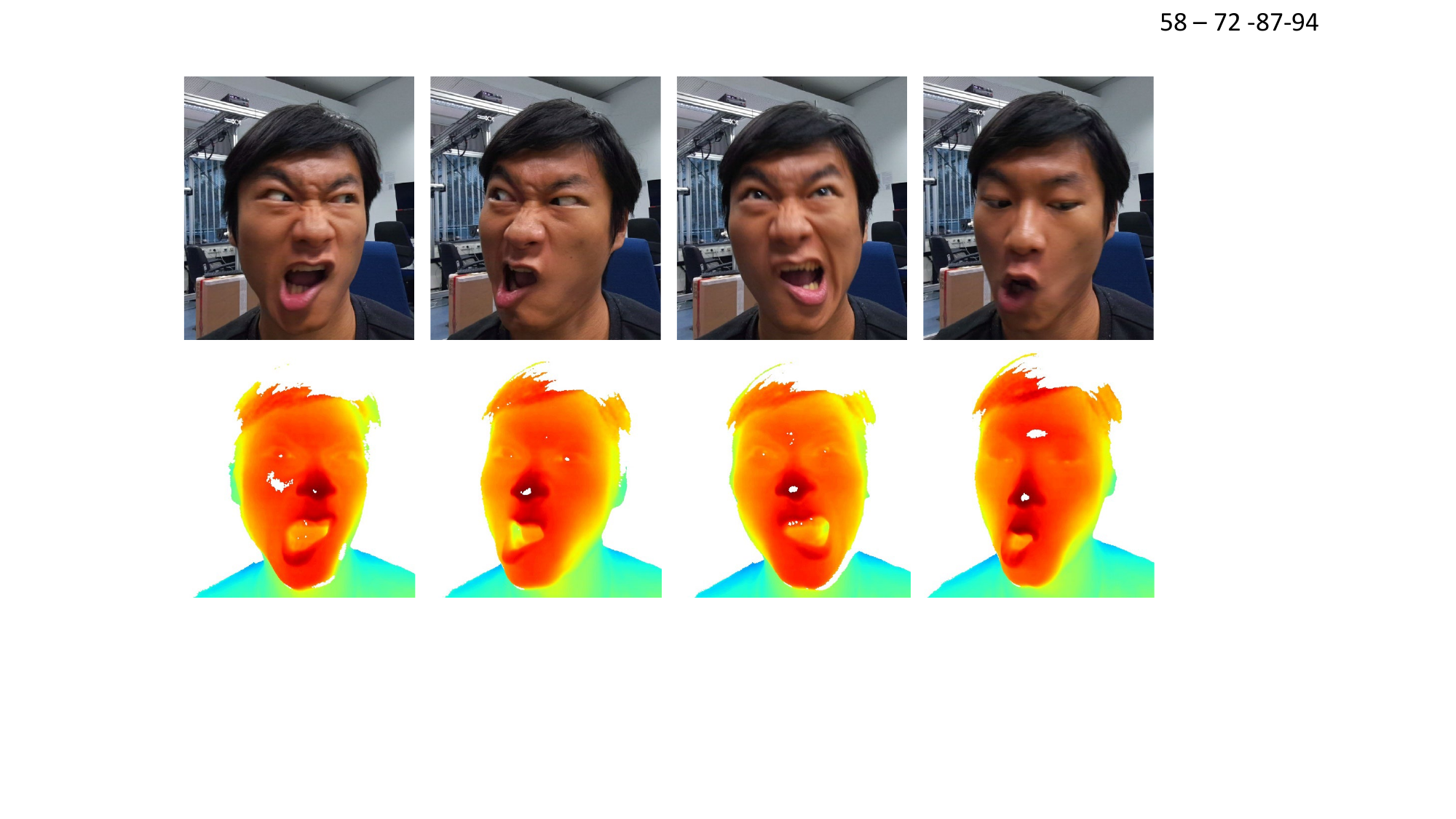}
    \caption{An example of captured DPHM-Kinect sequences with complex facial expressions and fast transitions.}
    \label{fig:kinect_dataset}
\vspace{-3mm}
\end{figure}

%
To assess head tracking performance, researchers commonly use datasets with rendered depth map sequences from talking face mesh sequences, such as VOCA~\cite{VOCA2019}. However, these sequences often capture limited lip movements during speech without complex, intricate facial expressions. In our work, we establish a challenging benchmark for head tracking and reconstruction. 
We construct a benchmark containing 130 single-view depth scan sequences, capturing diverse facial expressions in motion sequences, including rapid transitions. 
Specifically, we use a Microsoft Kinect Azure RGB-D camera to record data. The dataset includes eighteen men and eight women from different skin tones and ethnicities. For each participant, we recorded five sequences, where they quickly switched their expressions and could have global rotations, including 'smile and laugh,' 'eyeblinks,' 'fast-talking,' 'random facial expressions,' and 'mouth movements.' An example of the collected RGBD sequences is depicted in Fig.~\ref{fig:kinect_dataset}.
\begin{figure*}[!htp]
    \vspace{-5mm}
    \centering
    \includegraphics[width=0.95\linewidth]{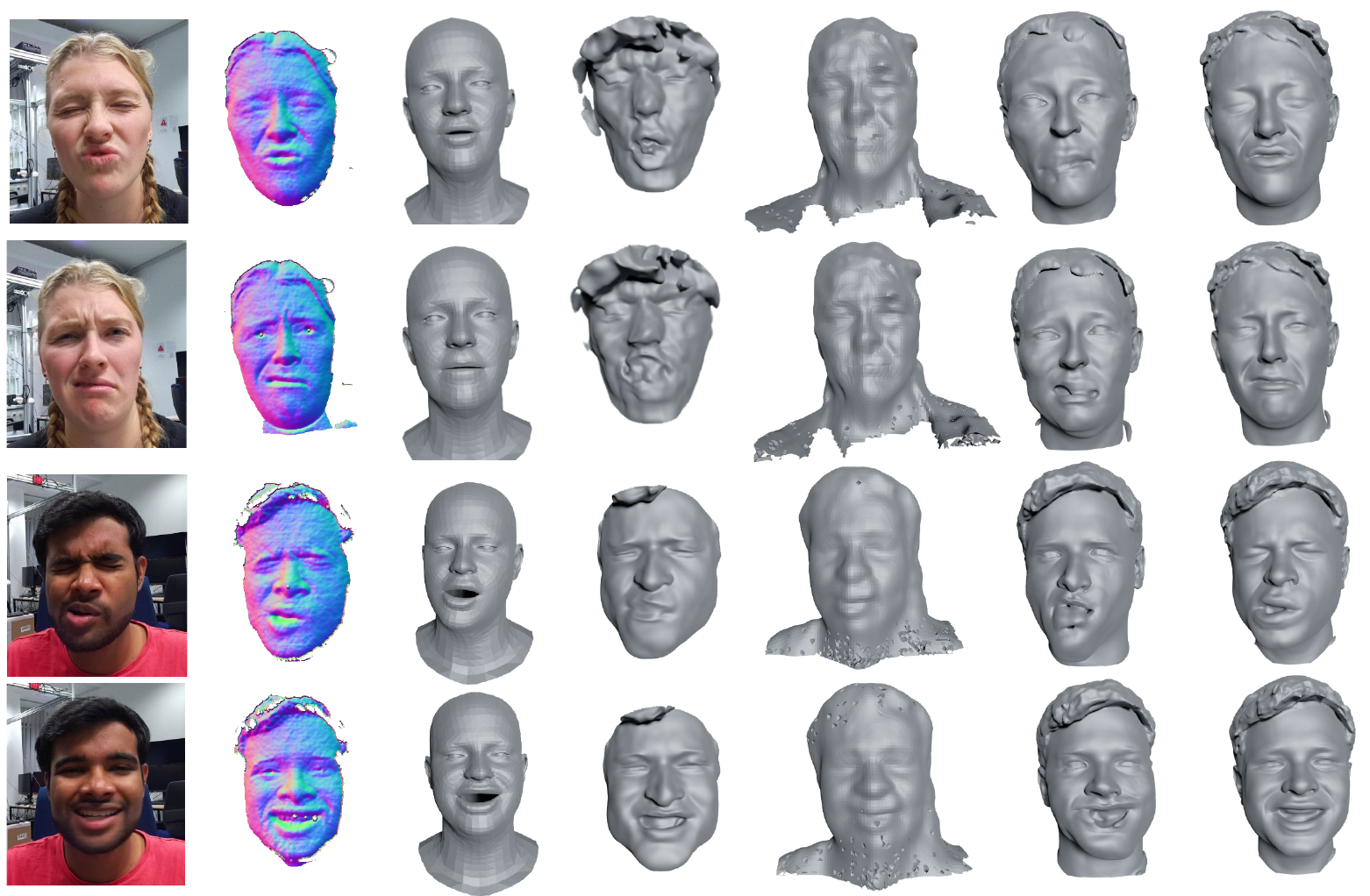}
    \begin{tabular}{p{15pt}p{40pt}p{60pt}p{55pt}p{60pt}p{55pt}p{70pt}p{70pt}p{70pt}}
                & (a) RGB 
                & (b) Input Scans
                & (c) FLAME
                & (d) ImFace*
                & (e) ImAvatar 
                & \quad (f) NPHM
                & (g) Ours
    \end{tabular}
    \caption{Head Tracking on the DPHM-Kinect dataset. Note that RGB images are only used for reference not used by all the methods except ImAvatar. Compared to state-of-the-art methods, our approach achieves more accurate identity reconstruction with detailed hair geometries while tracking more plausible expressions, even during extreme mouth movements.}
    \label{fig:kinect_tracking}
    \vspace{-4mm}
\end{figure*}
\section{Experiments}
\paragraph{Dataset}
To learn high-quality head geometry priors, we use 4,760 high-fidelity head scans with varying expressions of 239 identities from NPHM as the training dataset.
To evaluate the performance of head reconstruction and tracking from monocular scans,  we additionally 
use the reconstructed single-view depth sequences from a multi-view video dataset~\cite{kirschstein2023nersemble} by COLMAP~\cite{schoenberger2016mvs, schoenberger2016sfm}. We use twenty head motion sequences from 10 different identities.

\paragraph{Baselines} 
We compare against state-of-the-art head tracking methods: FLAME~\cite{li2017learning}, ImFace~\cite{zheng2022imface}, ImAvatar~\cite{zheng2022imavatar}, and NPHM~\cite{giebenhain2023learning}.
FLAME is the most recent and advanced template-based PCA model.
ImFace~\cite{zheng2022imface} is one of the pioneering works to integrate neural signed distance and deformation fields for face reconstruction.  We also include ImFace trained on the NPHM dataset notated as ImFace*.
ImAvatar~\cite{zheng2022imavatar} is an optimization method that utilizes a coordinate-MLP to reconstruct the personalized canonical head geometry and animates head motions through skinning weight fields, building upon FLAME. We also include depth supervision in the optimization of ImAvatar.
To isolate the effects of our proposed parametric diffusion, we compare against the improved version of NPHM based on backward-deformation expression space.

\paragraph{Evaluation Metrics}
To evaluate the accuracy of the head reconstruction and tracking, we calculate the $\ell_2$ distance error between the observed scans and reconstructed meshes. Concretely, for each point in the captured scan, we find the closest point in the reconstructed surface mesh and then calculate $\ell_2$ distance (mm).  Based on the nearest neighborhood search correspondence, we calculate the normal consistency ($\NC$) through cosine similarity. Also, we provide Recall scores ($\RC$) with thresholds of  $\tau=1.5mm$ and $2\tau=3mm$, demonstrating the percentage of observed points well approximated under a defined threshold.

\paragraph{Implementations}
Our DPHM is trained on a single RTX A6000. Initially, we train the identity and expression decoders on the NPHM dataset to obtain over-parametrized latent representations. This training involves using a learning rate of $0.001$ with a batch size of 32 for 6,000 epochs. The learning rate is decayed by a factor of 2 every 500 epochs. Subsequently, we train the identity and expression parametric diffusion models, employing UNet-1D~\cite{ronneberger2015u} as the denoising network based on DDPM~\cite{ho2020denoising}. These diffusion models are individually trained using a batch size of 32 and a learning rate of $\expnumber{8}{-5}$ for 200,000 iterations.

The test-time optimization of head tracking involves three phases. We firstly optimize $\Vec{z}^{id}$ and $\Vec{z}^{ex}_1$ for the first frame. The optimization is performed for 200 iterations, with a learning rate of $0.01$ and a decay of 0.1 every 50 iterations. Subsequently, we fix $\Vec{z}^{id}$ and incrementally optimize $\Vec{z}^{ex}_{2:N}$ for subsequent frames. The optimization uses a learning rate of $0.001$ for 50 iterations, with a decay of 0.1 after 30 iterations. Finally, we jointly fine-tune $\Vec{z}^{id}$ and $\Vec{z}^{ex}_{1:N}$  with a learning rate of $0.0001$ for 20 iterations.
 The hyper-parameters in Equation~\ref{equaRecLoss} are set to $\lambda_{norm} = 0.025, \lambda_{id}=0.25, \lambda_{ex}=0.25, \lambda_{temp}=0.5$. At each iteration, the noisy latent $\hat{\Vec{z}}$ is generated by Eq.~\ref{Equadiffusion_repara} with $t \in [0.4, 0.6 ]$.
 Please refer to the supplementary material for more details.
\begin{figure*}[!htp]
    \vspace{-6mm}
    \centering
    \includegraphics[width=0.95\linewidth]{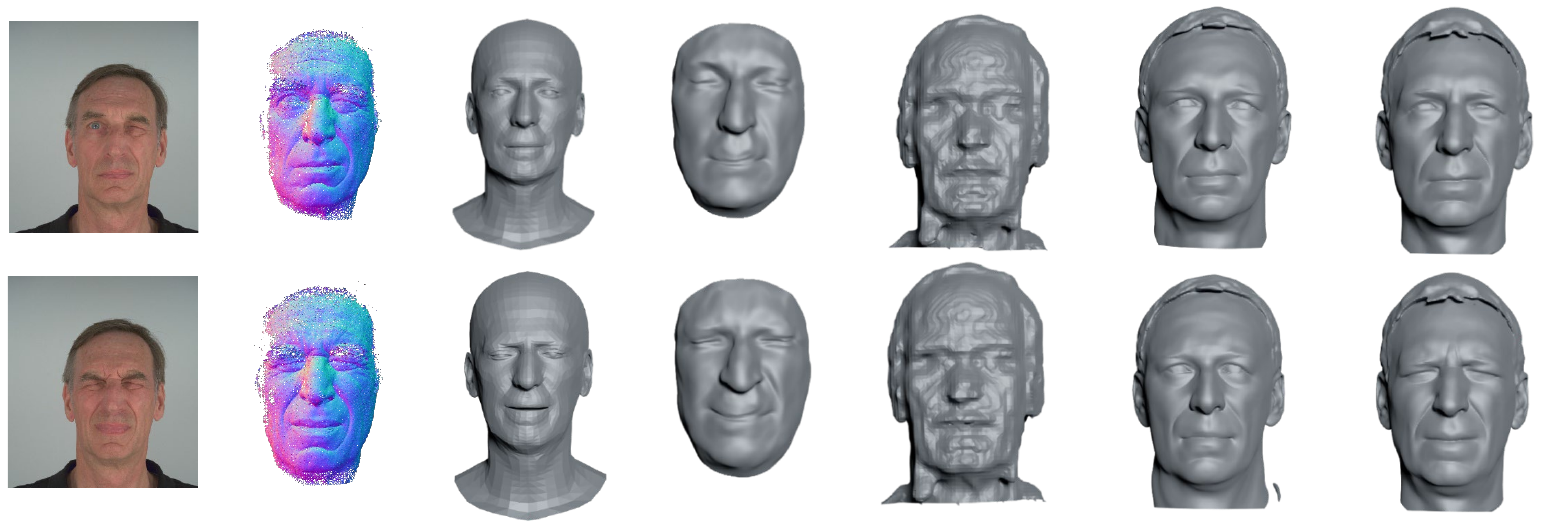}
        \begin{tabular}{p{10pt}p{50pt}p{60pt}p{55pt}p{60pt}p{55pt}p{60pt}p{60pt}p{60pt}}
                & (a) RGB 
                & (b) Input Scans
                & (c) FLAME
                & (d) ImFace*
                & (e) ImAvatar 
                &  (f) NPHM
                & (g) Ours
    \end{tabular}
    \vspace{-4mm}
    \caption{Head Reconstruction and Tracking on the single-view depth sequences of NerSemble~\cite{kirschstein2023nersemble}. Note that RGB images are only used for reference and not used by all methods except ImAvatar. Compared to state-of-the-art methods, our approach demonstrates the ability to reconstruct realistic head avatars with hairs and accurately capture intricate facial expressions such as eyelid movements.
    }
    \label{fig:mvs_tracking}
    \vspace{-2mm}
\end{figure*}
\subsection{Comparison against state of the art}
\paragraph{Head Tracking on DPHM-Kinect dataset.}
The qualitative comparison of our method with state-of-the-art reconstruction methods on the collected Kinect data is presented in Fig.~\ref{fig:kinect_tracking}.  FLAME struggles to model hair geometries and has a limited capacity for intricate facial expressions related to mouth and eye movements. ImFace focuses only on the front face region. ImAvatar always produces over-smooth results due to a lack of high-quality geometry priors. In contrast, NPHMs and our DPHMs successfully reconstruct fine-grained hair geometries by utilizing volumetric SDFs to model complete head geometries, accommodating different hairstyles. However, NPHMs often fail to reconstruct plausible expressions in challenging scenarios with partial and noisy scans. On the other hand, our DPHMs effectively regulate latent optimization, resulting in plausible reconstructions and accurate tracking, even in complicated expressions shown in the second and third rows. Quantitative comparisons in Table~\ref{tab:kinect} show that our approach consistently outperforms all state-of-the-art methods, illustrating more accurate and robust tracking.
\begin{table}[!htbp]
    \vspace{-1mm}
	\renewcommand\arraystretch{1.2}
    \setlength{\tabcolsep}{2pt}
	\begin{center}
        \footnotesize
		\begin{tabular}{*{10}{c}}
                \toprule
			Method & FLAME  & ImFace & ImFace* & IMAvatar & NPHM  & Ours\\ 
                \midrule
                $\ell_2$ $\downarrow$  &  5.251 & 11.21  & 11.00 & 5.581  & 1.579 & \textbf{1.465}\\
                NC $\uparrow$          &  83.88 & 84.82  & 84.00 & 70.23  & 85.97 & \textbf{86.80}\\
                RC $\uparrow$          &  15.80 & 57.70  & 58.74 & 11.30  & 65.95 & \textbf{70.79}\\
                RC2 $\uparrow$         &  53.11 & 71.88  & 72.99 & 40.75  & 88.33 & \textbf{90.98}\\
                \bottomrule
        \end{tabular}
          \vspace{-2mm}
        \caption{Quantitative comparison on the DPHM-Kinect dataset.} 
        \label{tab:kinect}
        \end{center}
    \vspace{-6mm}
\end{table}

\begin{table}[t]
        \vspace{-1mm}
	\renewcommand\arraystretch{1.2}
        \setlength{\tabcolsep}{2pt}
	\begin{center}
        \footnotesize
		\begin{tabular}{*{10}{c}}
                \toprule
			Method & FLAME  & ImFace & ImFace* & IMAvatar & NPHM  & Ours\\ 
                \midrule
                $\ell_2$   &  3.286   & 1.560  &  1.557  &  3.337 & 1.201 &  \textbf{1.103} \\
                NC           &  84.00  & 87.24 &  87.44   & 77.79  & 87.76 & \textbf{88.73} \\
                RC           &  21.06  & 80.57  & 80.60  & 14.97 & 84.73  & \textbf{89.96} \\
                RC2          &  60.70  & 92.69  & 93.16  & 54.26 & 96.63  & \textbf{97.20} \\
                \bottomrule
        \end{tabular}
       \vspace{-2mm}
        \caption{Quantitative comparison on dynamic point cloud sequences reconstructed from multi-view video dataset~\cite{kirschstein2023nersemble}. } 
        \label{tab:mvs}
        \end{center}
    \vspace{-6mm}
\end{table}

\paragraph{Head Tracking on NerSemble dataset }
Fig.~\ref{fig:mvs_tracking} depicts the comparisons on reconstructed single-view depth sequences from NerSemble~\cite{kirschstein2023nersemble}.
Again, we showcase the superiority of the diffusion prior-based regularization through improved numerical results across all metrics in Table~\ref{tab:mvs}. This is evident in the more accurate tracking of our results, particularly in terms of eye movements.

\subsection{Ablation Studies}
\begin{figure*}[!htp]
    \vspace{-5mm}
    \centering
    \includegraphics[width=0.90\linewidth]{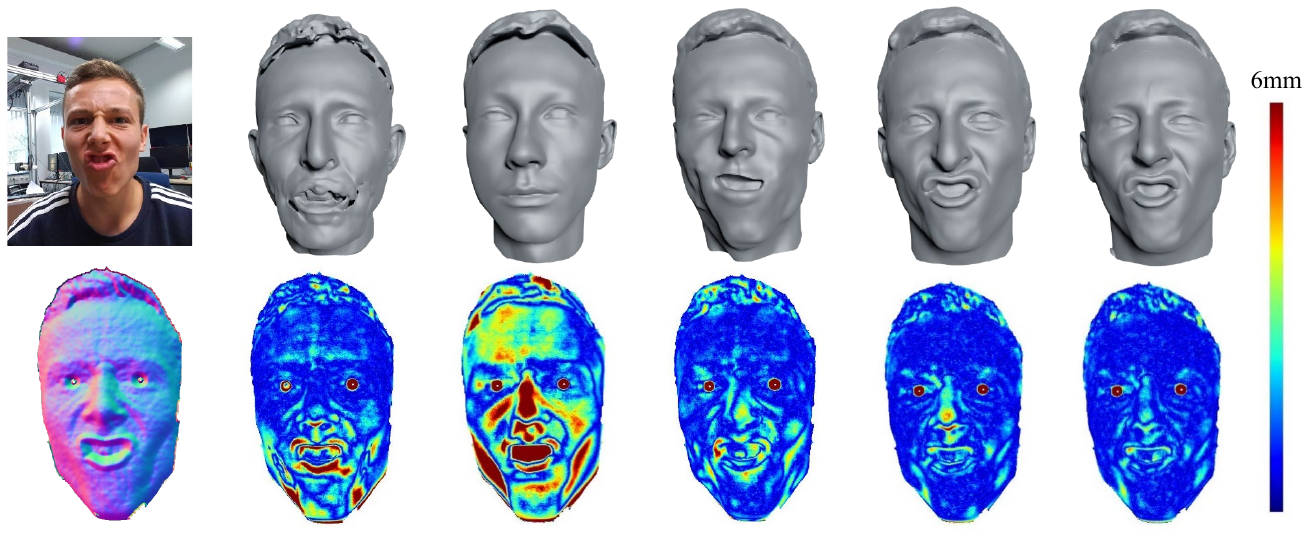}
    \begin{tabular}
        {p{35pt}p{70pt}p{65pt}p{55pt}p{60pt}p{60pt}p{60pt}}
                & (a)  
                & (b)  
                & (c)  
                & (d)  
                & (e)  
                & (f) 
    \end{tabular}
    \vspace{-2mm}
    \caption{ \textbf{Ablation Studies}
     (a) RGB reference \& Input Scans; (b) Ours with forward deformations; (c) Ours with VAE priors; (d)  Ours without expression diffusion; (e) Ours without identity diffusion;  (f) Ours.  Note that RGB images are only used for reference not used by all the methods except ImAvatar. We visualize the scan2mesh distance error map at the bottom.  Our final model captures complicated expressions with lower identity reconstruction errors. }
    \label{fig:ablation}
    \vspace{-1mm}
\end{figure*}

We conduct detailed ablation studies to verify the effectiveness of each design in our approach (see Table~\ref{tab:ablation} and Fig.~\ref{fig:ablation}).
\begin{table}[t]
    \vspace{-1mm}
	\renewcommand\arraystretch{1.2}
    \setlength{\tabcolsep}{2pt}
	\begin{center}
        \footnotesize
		\begin{tabular}{*{10}{c}}
                \toprule
			Method & forward  & VAE prior & w/o. exp. diff.  &  w/o. iden. diff   &  Ours  \\ 
                \midrule
                $\ell_2$ $\downarrow$  & 1.367  & 2.929  & 1.283 & 1.159    &  \textbf{1.146} \\
                NC $\uparrow$          & 80.58  & 82.82  & 90.60 & 91.13   &  \textbf{91.32} \\
                RC $\uparrow$          & 80.47  & 70.72  & 75.58 & 80.61   &  \textbf{81.33} \\
                RC2 $\uparrow$         & 93.20  & 68.85  & 94.23 & 95.93   &  \textbf{96.08} \\
                \bottomrule
        \end{tabular}
        \vspace{-2mm}
        \caption{Ablation studies on the DPHM-Kinect dataset.}
        \label{tab:ablation}
        \end{center}
        \vspace{-8mm}
\end{table}

\begin{figure*}
\vspace{-3mm}
    \centering
    \includegraphics[width=.45\linewidth]{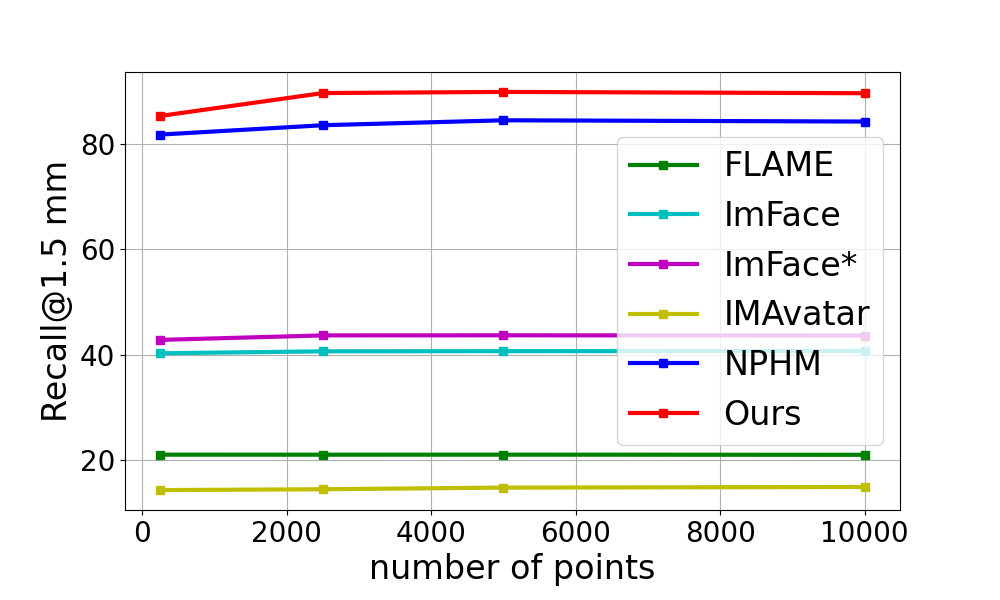}
    \hfill
    \includegraphics[width=.45\linewidth]{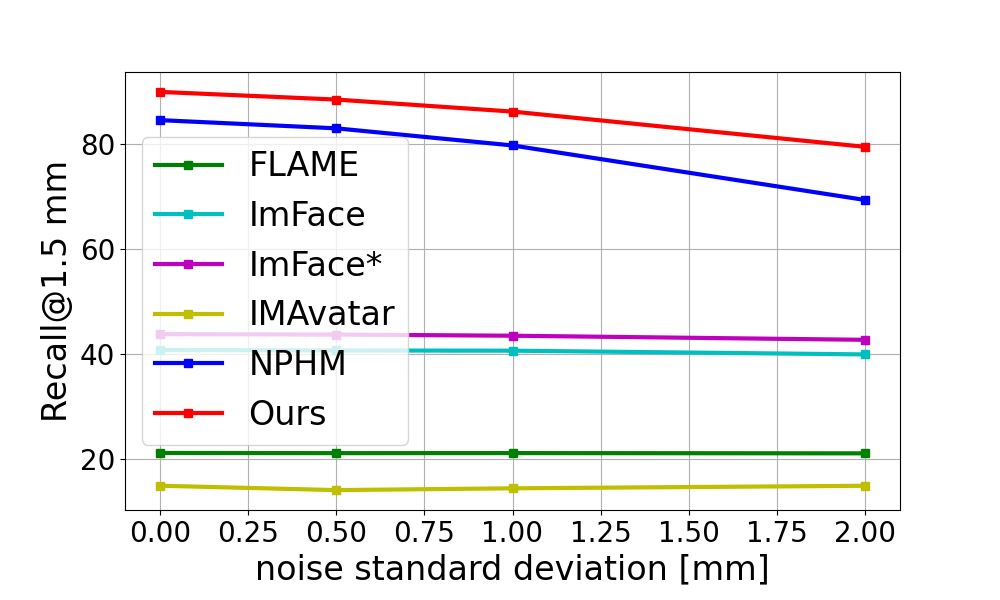}
    \vspace{-2mm}
    \caption{Robustness analysis with varying sparsity levels and additive Gaussian noise intensities. DPHMs demonstrate superior robustness compared to NPHMs across diverse imperfect observations}
    \label{fig:robust}
\vspace{-3mm}
\end{figure*}

\paragraph{What is the effect of backward deformation networks? }
An alternative approach to modeling facial expressions is to use forward deformation networks, as employed in the original NPHMs, which warp points from canonical space to expression space. Since all frames share the same mesh topology as in canonical space, it has a limited capacity to track complicated expressions. They always fail to reconstruct facial expressions with an open mouth.

\paragraph{What is the effect of the diffusion regularizer? }
We also explore using other regularizers to constrain the latent optimization. In line with V-Poser~\cite{SMPL-X:2019}, we trained a variational autoencoder (VAE)~\cite{kingma2013auto} using the over-parametrized latent and attempted to constrain the latent optimization using the VAE prior. However, we observed that the VAE priors cannot ensure plausible head identity reconstruction and fail to track correct expressions. This highlights the superiority of our diffusion prior-based regularizer.
%
%
%
%

\paragraph{What is the effect of the expression diffusion regularizer?}
We replace the expression diffusion regularizer with the simple one used in NPHM, which constrains the expression around the neutral state. Due to the weaker constraints, it always falls outside of the underlying surface manifold, leading to incorrect expression reconstruction. The obvious improvement in terms of numerical results can also verify the effective constraint of expression diffusion regularizer.

\paragraph{What is the effect of the identity diffusion regularizer?}
We replace the identity diffusion regularizer with the simpler one used in NPHM, which constrains the identity geometry to be close to the mean head in the training set. However, we observed that this simplified regularizer could not accurately reconstruct facial geometry details, particularly resulting in high errors in the nose part.

\subsection{Robustness Analysis}
To assess the robustness of our diffusion prior regularizations across various imperfect observations, we apply DPHMs to downsampled scans from NerSemble at sparsity levels of 10,000, 5,000, 2,500, and 250 points. 
We also evaluate varying intensities of additive Gaussian noise (standard deviations of 0mm, 0.5mm, 1mm, and 2mm) to sparse point clouds of size 5,000. Fig.~\ref{fig:robust} illustrates the Recall@$1.5mm$ curve under different sparsity and noise levels. Our method consistently outperforms all state-of-the-arts, demonstrating the superior robustness of our method.

\subsection{Limitations}

Despite achieving superior results, our approach currently has a limitation of slower inference due to the test-time optimization of neural parametric models. In the future, we will focus on real-time head-tracking solutions.

\section{Conclusion}
We introduced Diffusion Parametric Head Models (DPHMs), the first diffusion generative model enabling robust head reconstruction and tracking from real-world single-view depth sequences. Leveraging the diffusion priors of DPHMs, we designed a novel regularizer that effectively constrains the identity and expression codes on the underlying latent manifolds.
%
%
Extensive experiments and comparisons against state-of-the-art head reconstruction methods on a new challenging benchmark demonstrate that our method can reconstruct more accurate head geometries and achieve more robust and coherent expression tracking.
%

\paragraph{Acknowledgement.}
This work was supported by the ERC Starting Grant Scan2CAD (804724), the German Research Foundation (DFG) Research Unit “Learning and Simulation in Visual Computing, the ERC Starting Grant SpatialSem (101076253), as well as Sony Semiconductor Solutions. We thank Simon Giebenhain for close discussions and constructive suggestions during the project.
{
    \small
    \bibliographystyle{ieeenat_fullname}
    \bibliography{main}
}


\clearpage
\appendix
\section*{Appendix}
In this supplementary material, we delve into additional details about the network architectures in Sec.~\ref{SecNet}. Subsequently, we elaborate on collecting the DPHM-Kinect dataset in Sec.~\ref{SecDPHMKinect}. Following that, we provide a comprehensive explanation for the implementation of DPHMs for depth-based tracking in Sec.~\ref{SecImple}.   Moving forward, we showcase the results of unconditional head generation in Sec.~\ref{SecUncondGen}. 
%
Finally, we present supplementary comparisons against state-of-the-art head tracking methods in Sec.~\ref{SecAddCompare}, detailed results of the robustness analysis in Sec.~\ref{SecRobust}, along with some further discussions in Sec.~\ref{SecDiscuss}.
%
\section{Network Architectures}
\label{SecNet}
\subsection{Modified NPHMs}
\begin{figure}[t]
    \centering
    \includegraphics[width=\linewidth]{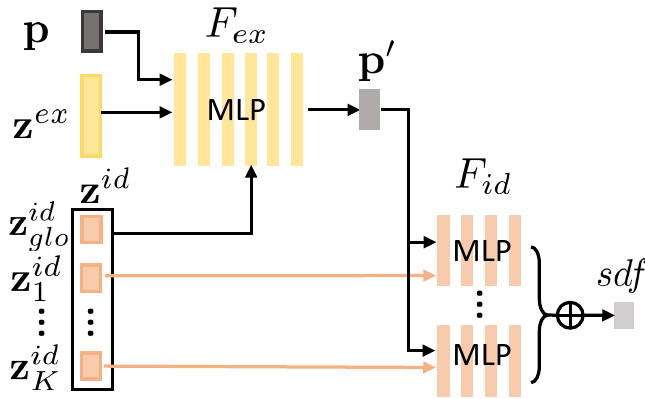}
    \caption{The network architecture of \textbf{our modified Neural Parametric Head Models} based on backward deformations.}
    \label{fig:network_nphm}
\end{figure}

In our DPHMs, we utilized a modified version of the Neural Parametric Head Models (NPHMs)~\cite{giebenhain2023learning,giebenhain2024mononphm} to learn over-parametrized latents from high-resolution head scans in the NPHMs dataset~\cite{giebenhain2023learning}. Specifically, we replaced the forward deformation network with the backward deformation network, enabling topology changes during expression tracking. The network architecture of the modified Neural Parametric Head Models is illustrated in Fig.~\ref{fig:network_nphm}.

We represent the human head geometry by a volumetric signed distance field decoded from two disentangled latents: the identity latent $\Vec{z}^{id}$ and the expression latent $\Vec{z}_{ex}$.  The $\Vec{z}^{id}$ is the concatenation of a global latent $\Vec{z}^{id}_{glo}$  and local latents $\Vec{z}^{id}_1, ..., \Vec{z}^{id}_K$.  $\Vec{z}^{id}_{glo}$ is used to estimate $K=39$ pre-defined anchor positions on a human head through a small $\MLP_{\anc}$. Each local identity latent $\Vec{z}^{id}_k$ is used to describe the head geometries around the $k_{th}$ anchor. To be specific, we define $K = 2 K_{\sym} + K_{\usym}$ facial anchors, denoted as $\Vec{a} \in \mathbb{R}^{K \times 3}$.
$K_{\sym}$ anchors are on the left face, mirrored to form the other $K_{\sym}$ anchors. $K_{\usym}$ anchors are positioned in the middle of the face, shared by both the left and right sides. 
To predict the SDF value of a point $\Vec{p} \in \mathbb{R}^3$ within the expression space, we concatenate it with the identity latent $\Vec{z}^{id}$ and expression latent  $\Vec{z}^{ex}$. The resulting feature is then passed through the backward deformation decoder $F_{ex}$, which warps $\Vec{p}$ to $\Vec{p}' \in  \mathbb{R}^5$ in the canonical space. It is important to note that $\Vec{p}'$ includes two hyper-dimensions ~\cite{park2021hypernerf} to model topological changes under different expressions, as continuous deformation fields alone may struggle with such changes.
Then, we feed $\Vec{z}^{id}$ and $\Vec{p}'$ into the canonical identity decoder $F_{id}$ to obtain its SDF prediction. The $F_{id}$ is implemented using an ensemble of smaller local Multi-Layer Perceptron (MLP)-based networks, each responsible for a local region centered around an anchor.
For the $k_{th}$ local region of facial anchor, we feed the corresponding local latent vector  $\Vec{z^{id}_{k}}$, along with the global latent vector $\Vec{z^{id}_{glo}}$ into an SDF decoder $\MLP_{\theta_k}$ with learnable weights $\Vec{\theta}_k$:
\begin{equation}
    f_k (\Vec{p}, \Vec{z^{id}_k}, \Vec{z^{id}_{glo}}) = \MLP_{\theta_k} ([\Vec{p} - \Vec{a}_k, \Vec{z^{id}_k}, \Vec{z^{glo}} ]) .
\end{equation}
To exploit facial symmetry, we share the network parameters and mirror the coordinates for each pair $(k, k^*)$ of symmetric regions:
\begin{equation}
    f_{k^*} (\Vec{p}, \Vec{z^{id}_{k^*}}, \Vec{z^{id}_{glo}}) = \MLP_{\theta_k} ([\flip(\Vec{p} - \Vec{a}_k), \Vec{z^{id}_k}, \Vec{z^{glo}} ]) .
\end{equation}
Finally, we can composite all local fields $f_k$ into a global field:
\begin{equation}
    F_{id} (\Vec{p}) = \sum_{k=1}^{K} \omega_k(\Vec{p}, \Vec{a_k}) f_k (\Vec{p}, \Vec{z^{id}_k}, \Vec{z^{id}_{glo}})  .
\end{equation}
The blending weights are calculated by a Gaussian kernel based on the Euclidean distance between the query point $\Vec{p}$ and $\Vec{a_k}$.
\[
    \omega^{'}_k(\Vec{p}, \Vec{a_k})= 
\begin{cases}
    \exp{\frac{- \| \Vec{p} - \Vec{a_k} \|_2}{2\sigma}} , & k >0 \\
    c,              & k=0
\end{cases}
\]

\begin{equation}
    \omega_k(\Vec{p}, \Vec{a_k}) = \frac{ \omega^{'}_k(\Vec{p}, \Vec{a_k}) } {\sum_{k'} \omega^{'}_k(\Vec{p}, \Vec{a_k})}
\end{equation}

We use a fixed isotropic kernel with standard deviation $\sigma$ and a constant response $c$ for $f_0$.
The global identity latent $\Vec{z^{id}_{glo}}$ has the dimension $d_{glo}=64$, and each local identity latent $\Vec{z^{id}_{k}}$ is of dimension $d_{loc}=32$. Therefore, the total dimension of $\Vec{z^{id}}$ is $(K+1) * d_{loc} + d_{glo} = 1344$. The backward deformation decoder $F_{ex}$ is implemented by a six-layer MLP with a hidden dimension of 512.  Each $\MLP_{\theta_k}$ for local SDF field prediction has four layers with a hidden size of 200. The anchor prediction MLP has three layers with a hidden size of 128. To blend the ensemble of local SDF fields, we use $\sigma=0.1$ and $c=\exp^{-0.2/{\sigma^2}}$.

\subsection{Denoiser networks.}
\begin{figure}[!htp]
    \centering
    \includegraphics[width=\linewidth]{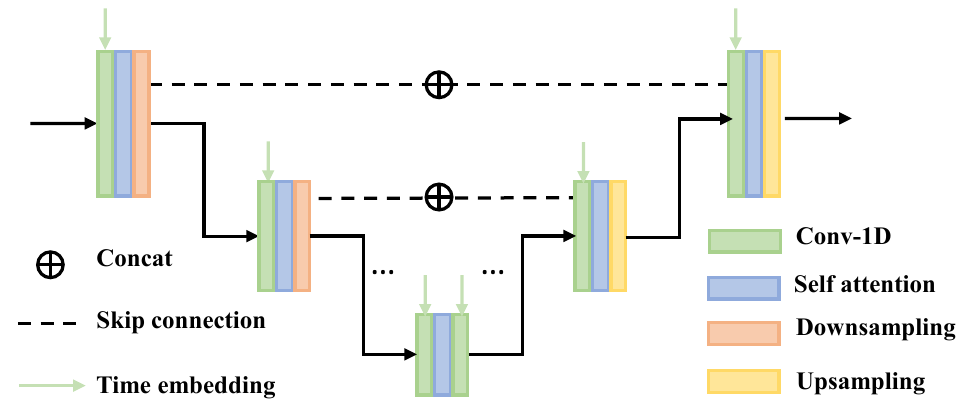}
    \caption{The denoiser network of our identity and expression latent diffusion models.}
    \label{fig:network_diff}
\end{figure}
The identity and expression diffusion models are constructed using UNet-1D~\cite{ronneberger2015u} architecture with incorporated attention layers~\cite{vaswani2017attention}, following DDPM~\cite{ho2020denoising}.
We analogously treat the identity and expression latents as sequences of 1D scalar features, with the only difference being the sequence length, which is equal to the latent feature dimension. Fig.~\ref{fig:network_diff} illustrates the detailed network architecture, which has an encoder of 4 downsampling blocks and a decoder of 4 upsampling blocks. 
The encoder progressively increases the feature dimension from 1 to 64, 128, 256, and 512, while simultaneously reducing the sequence length by 2.  The decoder follows the opposite pattern, reducing the feature dimension and doubling the sequence length.

\section{DPHM-Kinect dataset}
\label{SecDPHMKinect}
\begin{figure*}[!htp]
    \vspace{-5mm}
    \centering
        \begin{tabular}{cl}
              \multicolumn{1}{c}{} & \multirow{52}{*}{\includegraphics[width=0.62\linewidth]{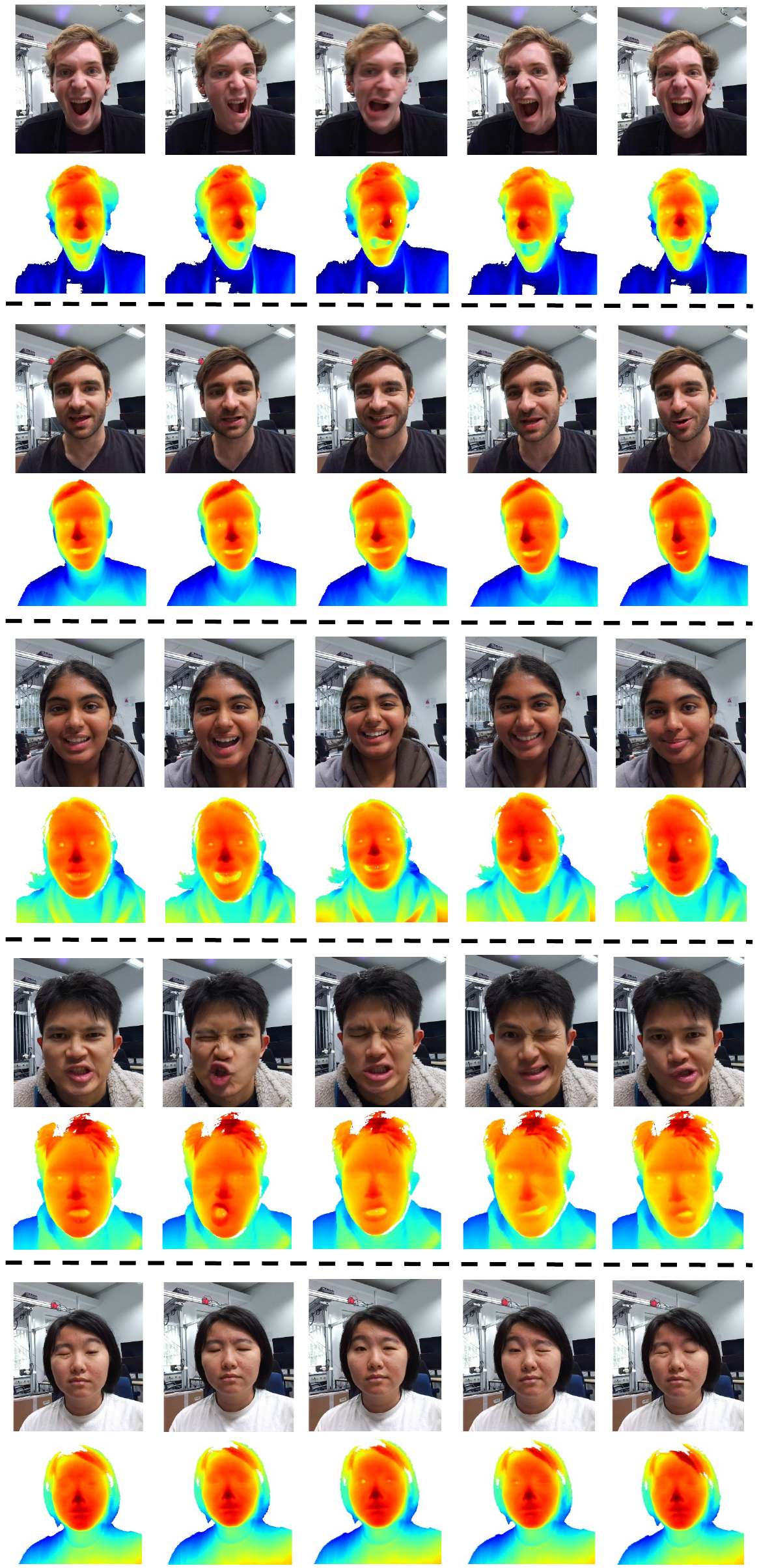}} \\
              \multicolumn{1}{c}{} & \\
               \multicolumn{1}{c}{} & \\
               \multicolumn{1}{c}{} & \\
               \multicolumn{1}{c}{} & \\
                \multicolumn{1}{c}{"Mouth movements"} & \\
                \multicolumn{1}{c}{} & \\
              \multicolumn{1}{c}{} & \\
              \multicolumn{1}{c}{} & \\
              \multicolumn{1}{c}{} & \\
               \multicolumn{1}{c}{} & \\
                \multicolumn{1}{c}{} & \\
               \multicolumn{1}{c}{} & \\
               \multicolumn{1}{c}{} & \\
               \multicolumn{1}{c}{} & \\
               \multicolumn{1}{c}{} & \\
                \multicolumn{1}{c}{"Fast talking"} & \\
              \multicolumn{1}{c}{} & \\
               \multicolumn{1}{c}{} & \\
               \multicolumn{1}{c}{} & \\
               \multicolumn{1}{c}{} & \\
                \multicolumn{1}{c}{} & \\
               \multicolumn{1}{c}{} & \\
               \multicolumn{1}{c}{} & \\
               \multicolumn{1}{c}{} & \\
               \multicolumn{1}{c}{} & \\
                \multicolumn{1}{c}{"Smile and laugh"} & \\
             \multicolumn{1}{c}{} & \\
               \multicolumn{1}{c}{} & \\
               \multicolumn{1}{c}{} & \\
               \multicolumn{1}{c}{} & \\
                \multicolumn{1}{c}{} & \\
               \multicolumn{1}{c}{} & \\
               \multicolumn{1}{c}{} & \\
               \multicolumn{1}{c}{} & \\
               \multicolumn{1}{c}{} & \\
                  \multicolumn{1}{c}{} & \\
                \multicolumn{1}{c}{"Random expressions"} & \\
              \multicolumn{1}{c}{} & \\
               \multicolumn{1}{c}{} & \\
               \multicolumn{1}{c}{} & \\
               \multicolumn{1}{c}{} & \\
                \multicolumn{1}{c}{} & \\
               \multicolumn{1}{c}{} & \\
               \multicolumn{1}{c}{} & \\
               \multicolumn{1}{c}{} & \\
               \multicolumn{1}{c}{} & \\
                \multicolumn{1}{c}{} & \\
                \multicolumn{1}{c}{"Eyeblinks"} & \\
               \multicolumn{1}{c}{} & \\
               \multicolumn{1}{c}{} & \\
               \multicolumn{1}{c}{} & \\
               \multicolumn{1}{c}{} & \\
    \end{tabular}
    \caption{ Example sequences of our \textbf{DPHM-Kinect dataset}.}
    \label{fig:kinect_dataset_supple}
\end{figure*}

The monocular RGBD sequences of our DPHM-Kinect dataset are collected from different skin tones and ethnicities. After obtaining consent from each attendee, we recorded five types of head motion sequences: ' smile and laugh,' 'eyeblinks,' 'fast-talking,' 'random facial expressions,' and 'mouth movements.' The recording framerate is 16$\text{fps}$. Each motion sequence lasts 6-10$\text{s}$, thus containing 96-160 frames. During data capture, they sit in front of the Kinect Azure sensor at a 15-40$\text{cm}$ distance. The examples of these motion types from different attendees are depicted in Fig.~\ref{fig:kinect_dataset_supple}. 

\section{Implementation Details}
\label{SecImple}
\paragraph{Proprocessing.}
We begin by eliminating background pixels from depth maps using a threshold of $d_{\text{max}}=0.6m$. Subsequently, bilateral smoothing is applied to the depth maps. Following this, surface normals are computed using the cross-product of derivatives along the x and y directions. Next, the depth and normal maps are lifted into 3D space, resulting in oriented partial point clouds. Finally, we crop out the points within the head region as the input.

\paragraph{Rigid registration.}
Prior to non-rigid tracking,  we need to obtain the rigid transformation parameters that convert the provided scan from the camera coordinate system to that of DPHMs. Since the coordinate system of DPHMs aligns with the FLAME space~\cite{li2017learning}, we opt to perform FLAME fitting. This includes the optimization of identity, expression, and pose parameters, as well as scale, rotation, and translation parameters.

\paragraph{Non-rigid tracking.}
With the rigid alignment serving as initialization, our method optimizes the identity and expression latents for depth-based head tracking, simultaneously fine-tuning the rigid parameters. Our non-rigid tracking comprises three phases: 'identity fitting', 'expression fitting', and 'joint finetuning'.
In the first phase of identity fitting, we optimize the identity latent $\Vec{z}^{id}$ and expression latent $\Vec{z}^{ex}_1$ for the first frame. In the expression fitting, we fix the identity latent and optimize the expression latent frame by frame. The optimization result $\Vec{z}^{ex}_{i-1}$ of the last frame is used as initialization for the expression latent of the next frame $\Vec{z}^{ex}_{i}$.
In the joint fine-tuning stage, we finetune the identity latent $\Vec{z}^{id}$, all expression latents $\Vec{z}^{ex}_{1:N}$, as well as the rigid transformation parameters for better alignments.
%
%
%
\section{Unconditional Generation of DPHMs}
\label{SecUncondGen}
\subsection{Identity Generation}
\begin{figure*}[!htp]
    \vspace{-5mm}
    \centering
    \includegraphics[width=.96\linewidth]{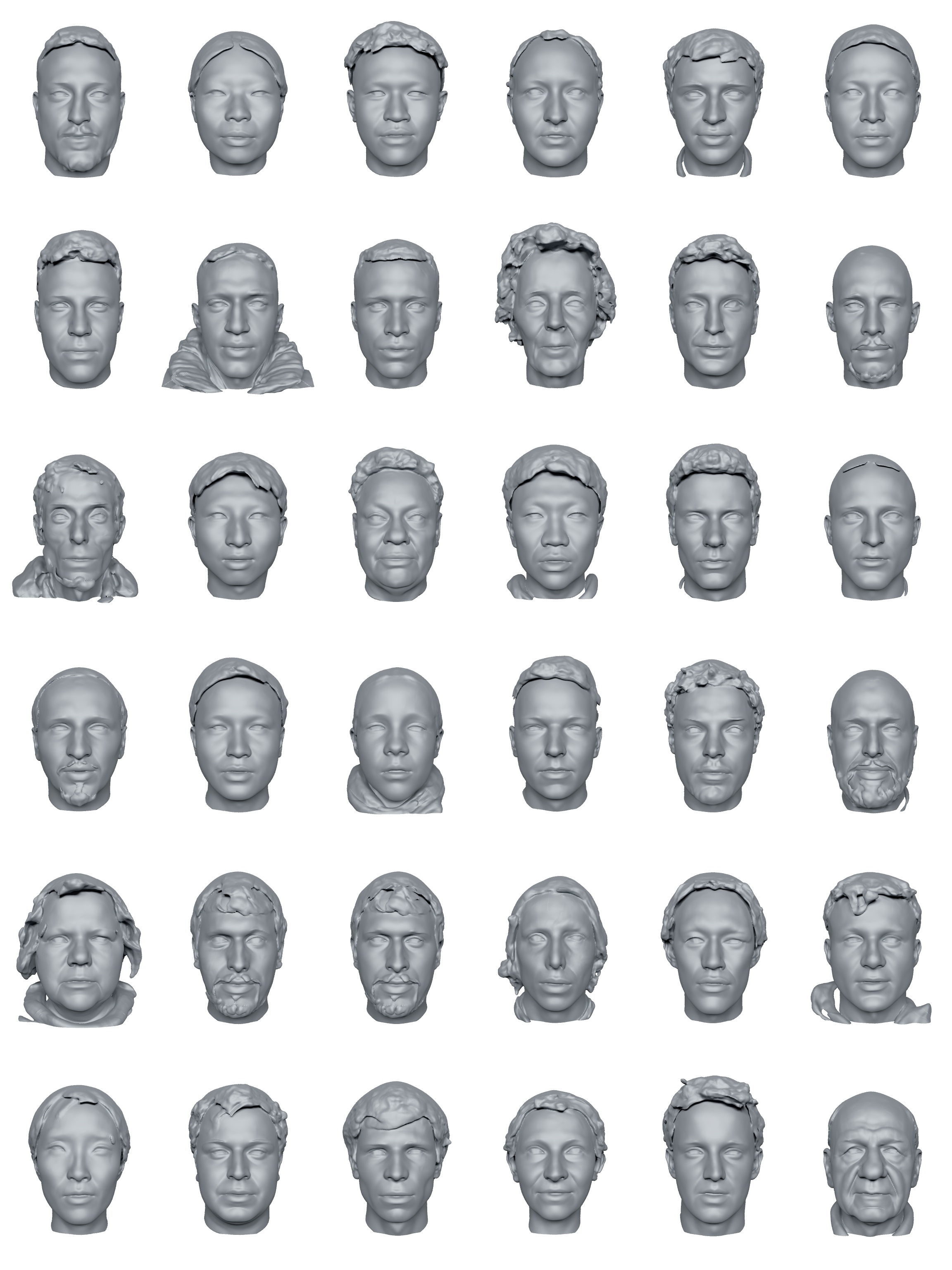}
    \caption{Unconditional sampling results of identity parametric diffusion model. Our approach can generate high-quality head avatars with diverse hairstyles.}
    \label{fig:iden_diff}
\end{figure*}

We present the randomly sampled results of our unconditional identity diffusion in Figure~\ref{fig:iden_diff}. Our approach demonstrates the ability to generate high-quality head avatars with diverse hairstyles.

\begin{figure*}[!htp]
    \vspace{-5mm}
    \centering
    \begin{subfigure}[t]{0.16\textwidth}
       \includegraphics[width=\linewidth]{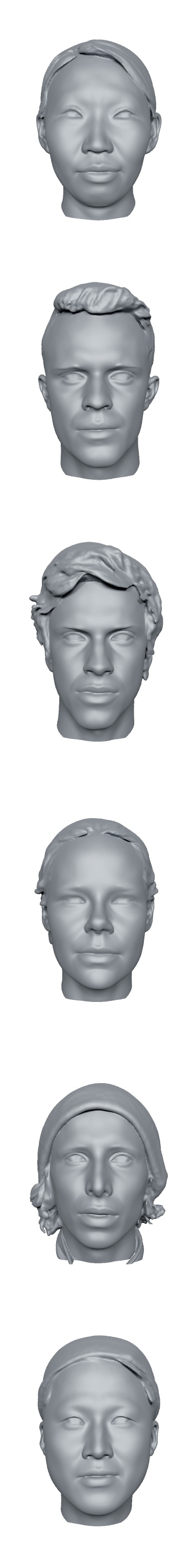}
    \end{subfigure}
    \rulesep
     \begin{subfigure}[t]{0.8\textwidth}
        \includegraphics[width=\linewidth]{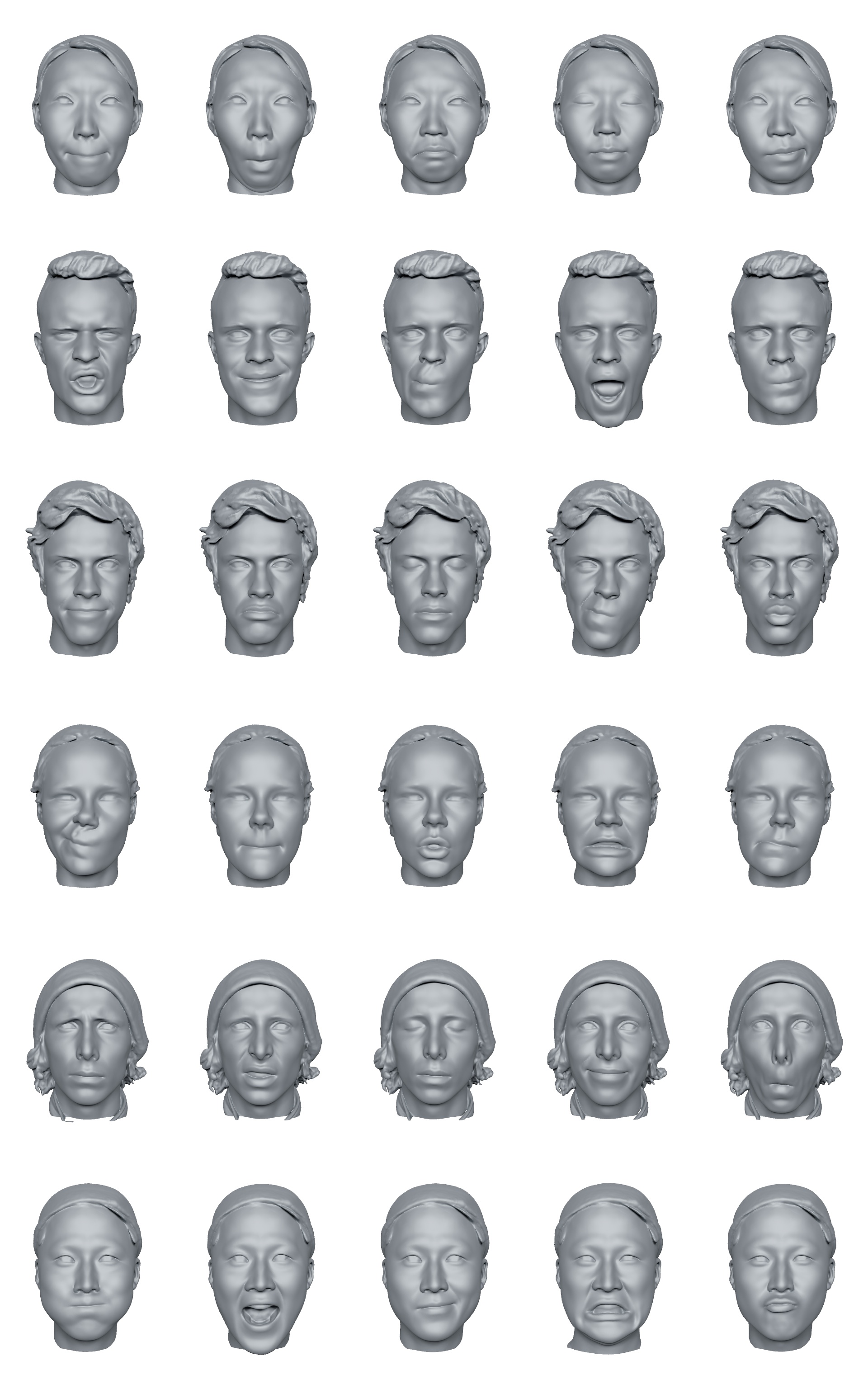}
    \end{subfigure}
    \caption{Unconditional sampling results of expression parametric diffusion model. The first column presents the canonical identity geometry with the neutral expression,~\ie, zero expression latent vector. Our method can generate a variety of plausible complex expressions.}
    \label{fig:expre_diff}
\end{figure*}
\subsection{Expression Generation}
In Figure~\ref{fig:expre_diff}, we present randomly sampled expressions for given head identities under the neutral expression with a closed mouth. Our expression parametric latent diffusion demonstrates the capability to generate various complicated facial expressions.

\section{Additional Comparisons}
\label{SecAddCompare}

\subsection{Additional comparisons on the DPHM-Kinect dataset}
\label{SubSecAddDPHM}
\begin{figure*}[!htp]
    \vspace{-8mm}
    \centering
    \includegraphics[width=0.94\linewidth]{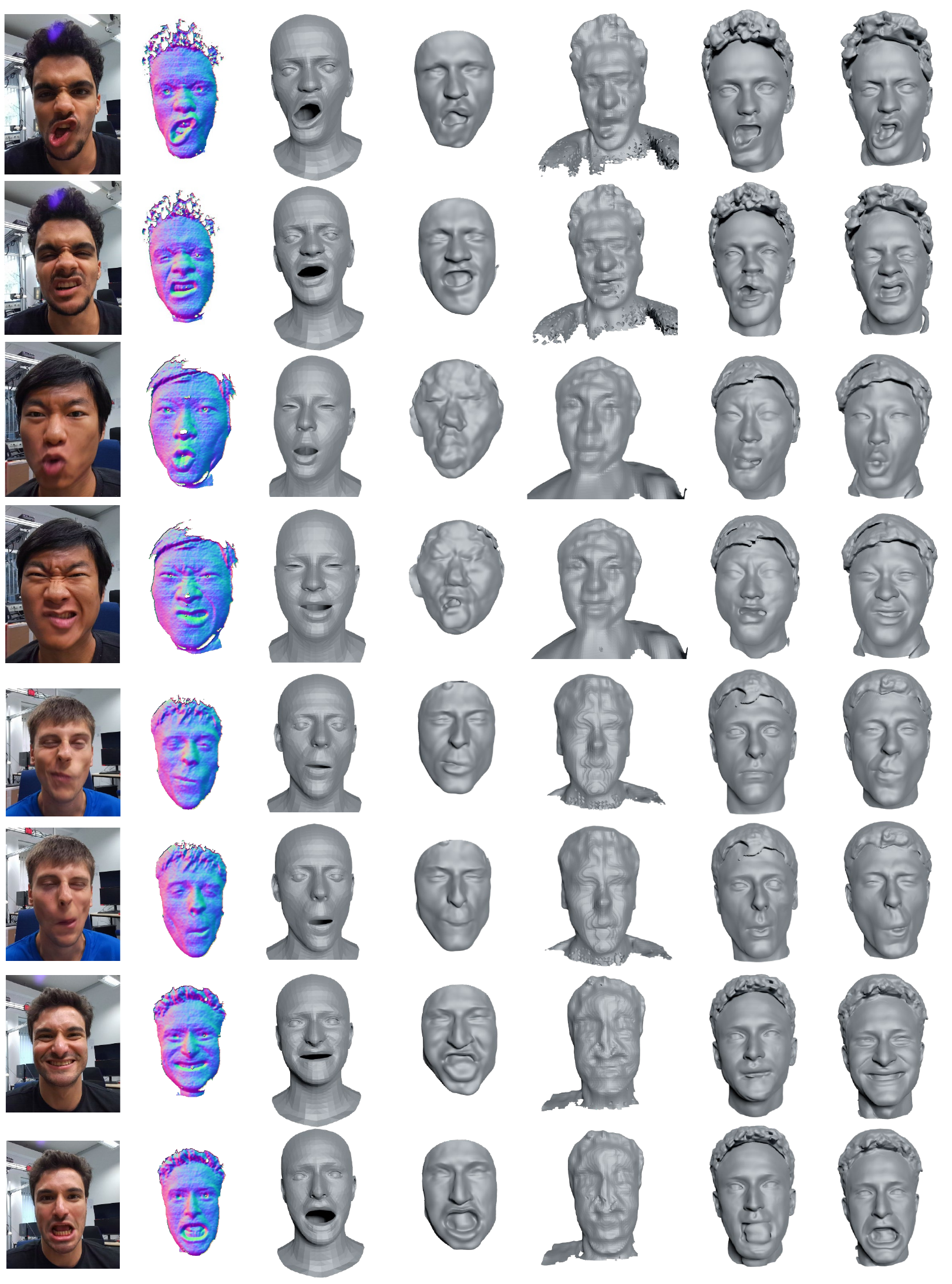}
    \begin{tabular}{p{15pt}p{40pt}p{60pt}p{55pt}p{60pt}p{55pt}p{70pt}p{70pt}p{70pt}}
                & (a) RGB 
                & (b) Input Scans
                & (c) FLAME
                & (d) ImFace*
                & (e) ImAvatar 
                & \quad (f) NPHM
                & (g) Ours
    \end{tabular}
    \vspace{-5mm}
    \caption{Head Tracking on the DPHM-Kinect dataset. Note that RGB images are only used for reference not used by all the methods except ImAvatar. Compared to state-of-the-art methods, our approach achieves more accurate identity reconstruction with detailed hair geometries while tracking more plausible expressions, even during extreme mouth movements.}
    \label{fig:kinect_tracking_supple}
\end{figure*}

%
In Figure~\ref{fig:kinect_tracking_supple}, we visualize additional comparisons on the monocular depth sequences from our DPHM-Kinect benchmark.

\subsection{Additional comparisons on the Multi-view Video dataset}
\label{SubSecAddMVS}
\begin{figure*}[!htp]
    \vspace{-5mm}
    \centering
    \includegraphics[width=0.95\linewidth]{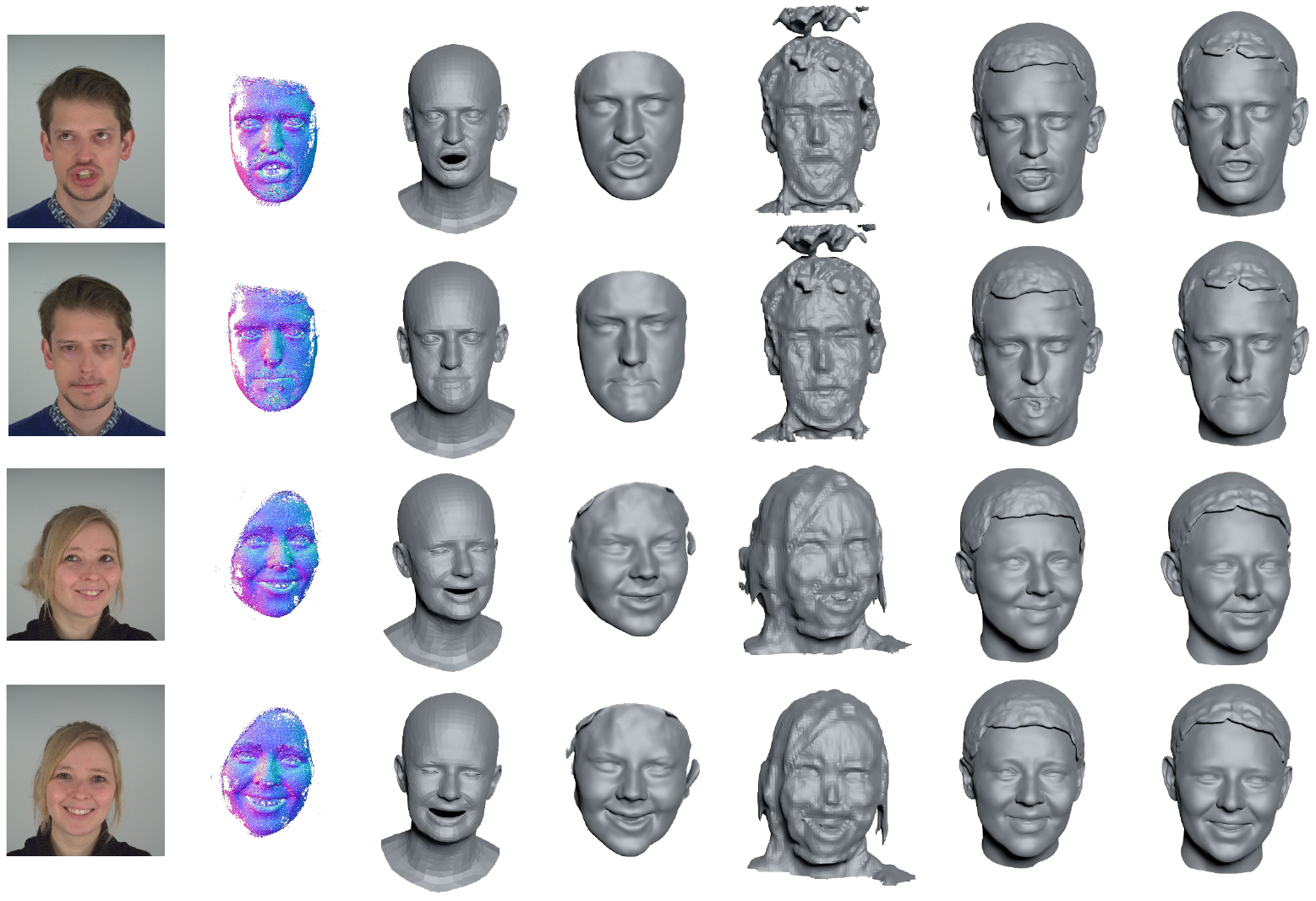}
        \begin{tabular}{p{8pt}p{50pt}p{60pt}p{55pt}p{55pt}p{55pt}p{65pt}p{60pt}p{60pt}}
                & (a) RGB 
                & (b) Input Scans
                & (c) FLAME
                & (d) ImFace*
                & (e) ImAvatar 
                &  (f) NPHM
                & (g) Ours
    \end{tabular}
    \caption{Head Reconstruction and Tracking on the single-view depth sequences of NerSemble~\cite{kirschstein2023nersemble}. Note that RGB images are only used for reference and not used by all methods except ImAvatar. Compared to state-of-the-art methods, our approach demonstrates the ability to reconstruct realistic head avatars with hairs and accurately capture intricate facial expressions.
    }
   \label{fig:mvs_tracking_supple}
\end{figure*}

In Figure~\ref{fig:mvs_tracking_supple}, we provide more qualitative comparisons on the single-view depth sequences reconstructed from the multi-view videos of NerSemble~\cite{kirschstein2023nersemble}.

\subsection{Evaluations on Unobserved Regions} 
During evaluations, we use partial depth scans used for test-time optimization as the target to calculate metrics.
For DPHM-Kinect sequences, we do not have the ground truth of dynamic head scans. Thus, we can only use the single-view Kinect depth scans for evaluation. 
To better evaluate the reconstruction of unobserved regions, we conduct additional comparisons on the NerSemble~\cite{kirschstein2023nersemble} dataset by only using single-view depth videos as input during optimization, while using more complete scans from multi-view depths as targets during evaluation. 
In Tab.~\ref{tab:oneview}, our method consistently outperforms existing methods in all metrics, which further confirms the effectiveness of our approach in reconstructing more accurate unobserved geometries.
%
\begin{table}[!htbp]
	\renewcommand\arraystretch{1.2}
        \setlength{\tabcolsep}{2pt}
	\begin{center}
        \footnotesize
            \begin{tabular}{*{10}{c}}
                \toprule
			Method & FLAME  & ImFace & ImFace* & ImAvatar  & NPHM  & Ours\\ 
                \midrule
                $\ell_2$     & 2.947  &  9.471  & 3.065   & 3.255  &  1.024 & \bf 0.894\\
                NC           & 69.50  &  78.74  & 84.36   & 78.13  &  88.07 & \bf 89.02\\
                RC           & 24.21  &  10.79  & 31.86   & 16.23  &  85.88 & \bf 91.02\\
                RC2          & 55.15  &  22.11  & 63.29   & 56.63  &  97.10 & \bf 97.99\\
                \bottomrule
        \end{tabular}
            \caption{Head tracking reconstructed from single-view depth scans. The results are evaluated on multi-view depth scans.} 
        \label{tab:oneview}
        \vspace{-4mm}
        \end{center}
\end{table}
\subsection{Comparison with Template-based Non-rigid Registration Method.} 
\begin{figure}[!htpb]
\centering
\includegraphics[width=\linewidth]{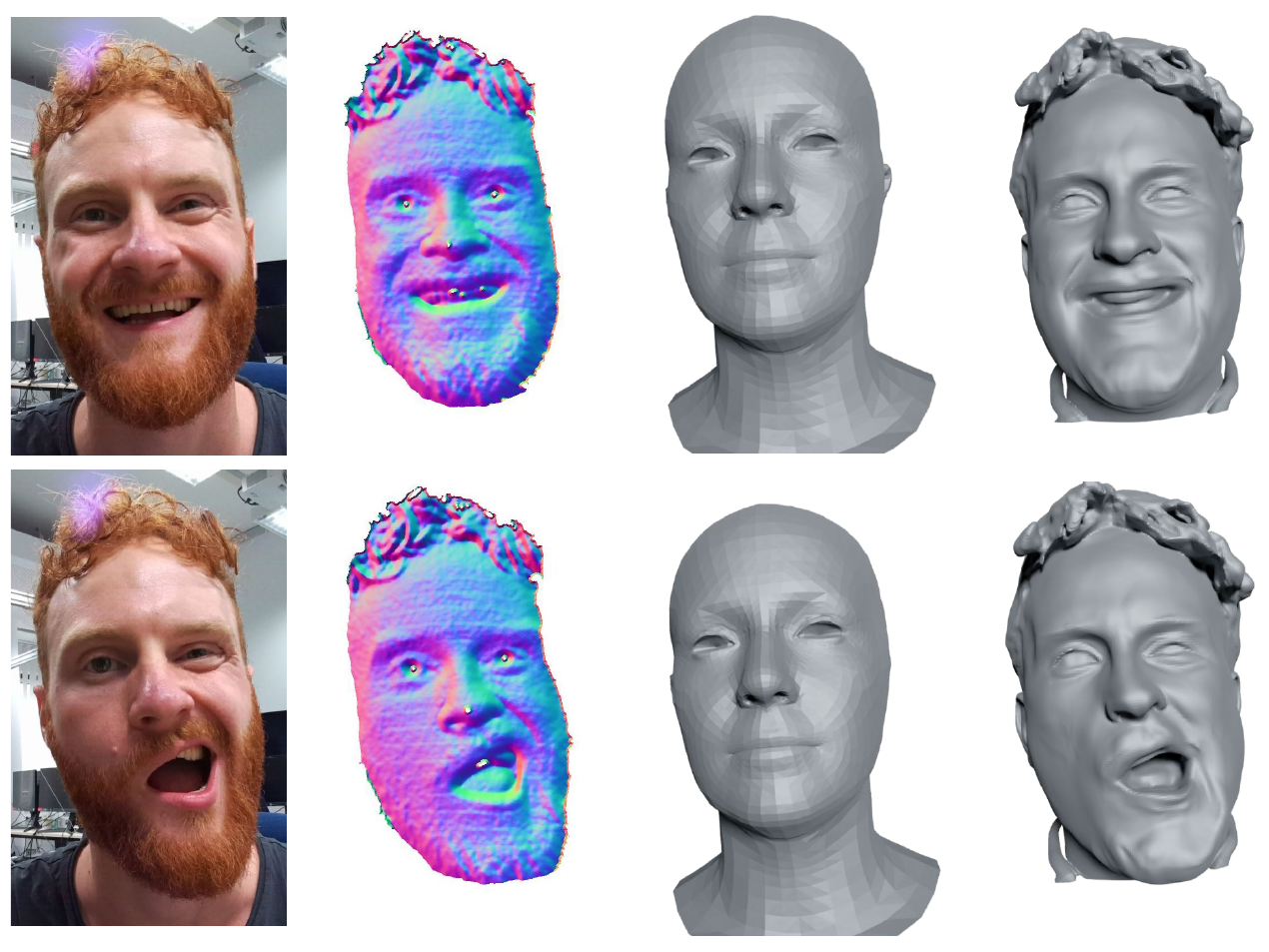}
\begin{tabular}{p{40pt}p{60pt}p{45pt}p{45pt}}
                 (a) RGB 
                & (b) Input Scans
                & (c) NICP
                & (d) Ours
    \end{tabular}
\caption{Qualitative comparisons against NICP on the DPHM-Kinect dataset.}
\label{fig:compare_nicp}
\end{figure}

Additionally, we include a classic template-based non-rigid registration method, NICP~\cite{amberg2007optimal}, into our comparisons. We re-implement NICP using a template mesh from FLAME~\cite{li2017learning} to obtain mesh deformations that align with depth scans.
We evaluate against NICP using the DPHM-Kinect dataset, which contains more challenging expressions compared to VOCA~\cite{VOCA2019}. As illustrated in Fig.~\ref{fig:compare_nicp},  NICP cannot recover plausible identities with hair and correct expressions because it does not have effective priors to handle imperfect observations of noisy and partial scans. 
It also cannot perform expression transfer, without the disentanglement of identity and expression. 
The quantitative comparisons presented in Tab.~\ref{tab:compare_nicp} demonstrate the superiority of our approach again.
%

\begin{table}[!htbp]
    \vspace{-2mm}
    \renewcommand\arraystretch{1.2}
    \setlength{\tabcolsep}{2.3pt}
	\begin{center}
		\begin{tabular}{*{5}{c}}
                \toprule
			Metric & $\ell_2$ $\downarrow$  & NC $\uparrow$  & RC $\uparrow$ &  RC2 $\uparrow$ \\ 
                \midrule
                NICP & 3.926 & 81.47  & 32.32 & 70.50 \\
                 Ours & \textbf{1.465} & \textbf{86.80}  & \textbf{70.79} & \textbf{90.98}\\
                \bottomrule
        \end{tabular}
        \caption{Quantitative comparisons against NICP on the DPHM-Kinect dataset.} 
        \label{tab:compare_nicp}
        \end{center}
        \vspace{-4mm}
\end{table}

\section{Robustness Analysis}
\label{SecRobust}
\begin{figure}[!htp]
    \centering
    \begin{tabular}{cl}
              \multicolumn{1}{c}{} & \multirow{10}{*}{\includegraphics[width=.8\linewidth]{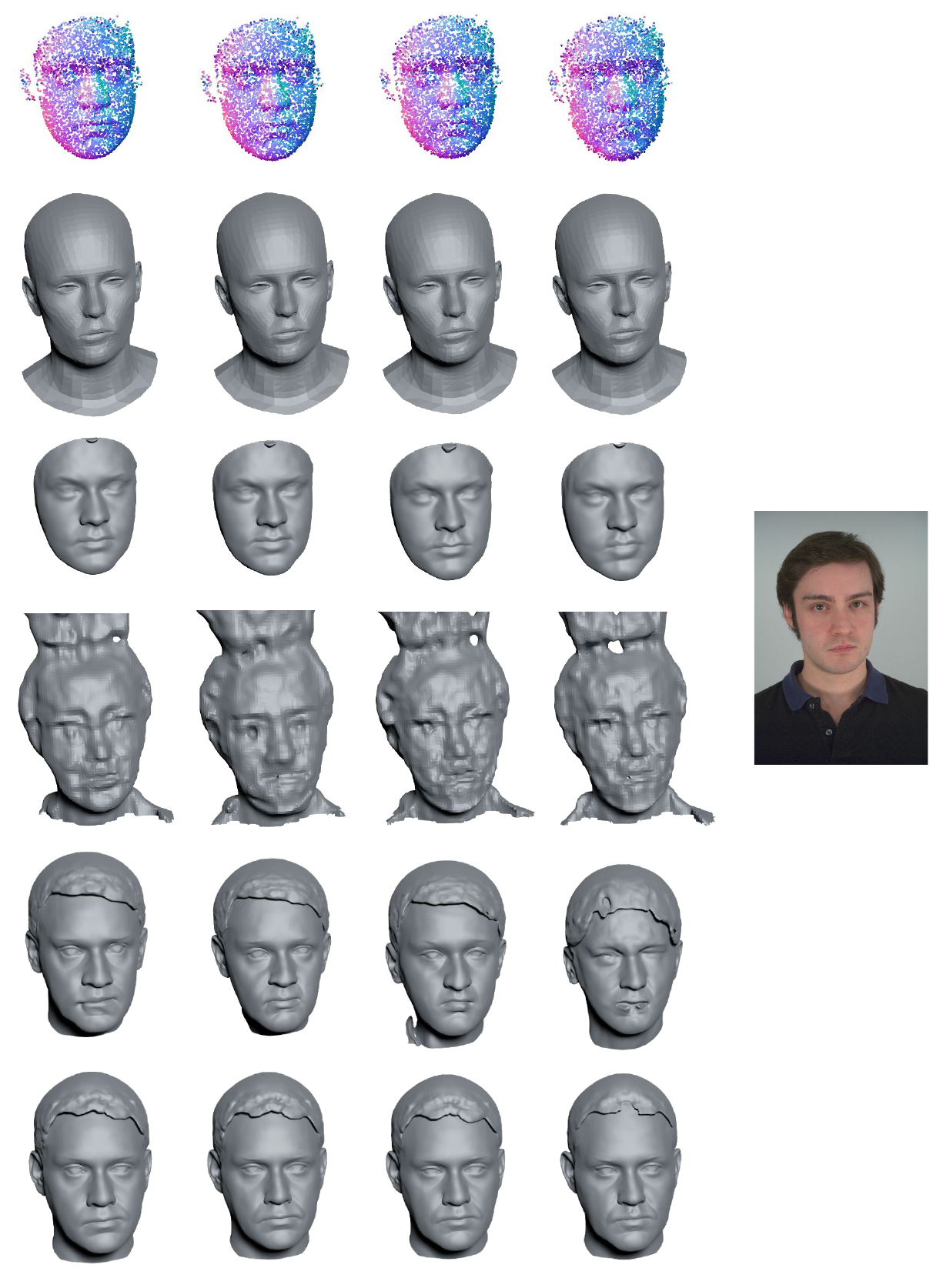}}\\
               \multicolumn{1}{c}{Input} & \\
             \multicolumn{1}{c}{Scans} & \\
              \multicolumn{1}{c}{} & \\
                \multicolumn{1}{c}{} & \\
                \multicolumn{1}{c}{FLAME} & \\
              \multicolumn{1}{c}{} & \\
                \multicolumn{1}{c}{} & \\
                \multicolumn{1}{c}{ImFace*} & \\
              \multicolumn{1}{c}{} & \\
                \multicolumn{1}{c}{} & \\
                \multicolumn{1}{c}{ImAvatar} & \\
              \multicolumn{1}{c}{} & \\
                \multicolumn{1}{c}{} & \\
                 \multicolumn{1}{c}{} & \\
                \multicolumn{1}{c}{NPHM} & \\
              \multicolumn{1}{c}{} & \\
              \multicolumn{1}{c}{} & \\
              \multicolumn{1}{c}{} & \\
                \multicolumn{1}{c}{Ours} & \\
              \multicolumn{1}{c}{} & \\
    \end{tabular}
    \begin{tabular}{p{38pt}p{20pt}p{20pt}p{20pt}p{25pt}p{25pt}}
                 Noise std
                & 0mm
                & 0.5mm
                &  1mm
                &  2mm
                &  RGB
    \end{tabular}
    \caption{Qualitative comparisons of NPHM and our method
with respect to noise in the input scan. We perturb the input scans by additive Gaussian noise with different standard deviations. }
    \label{fig:robust_noise_supple}
\end{figure}

\begin{figure}[!htp]
    \centering
        \begin{tabular}{cl}
              \multicolumn{1}{c}{} & \multirow{10}{*}{\includegraphics[width=.8\linewidth]{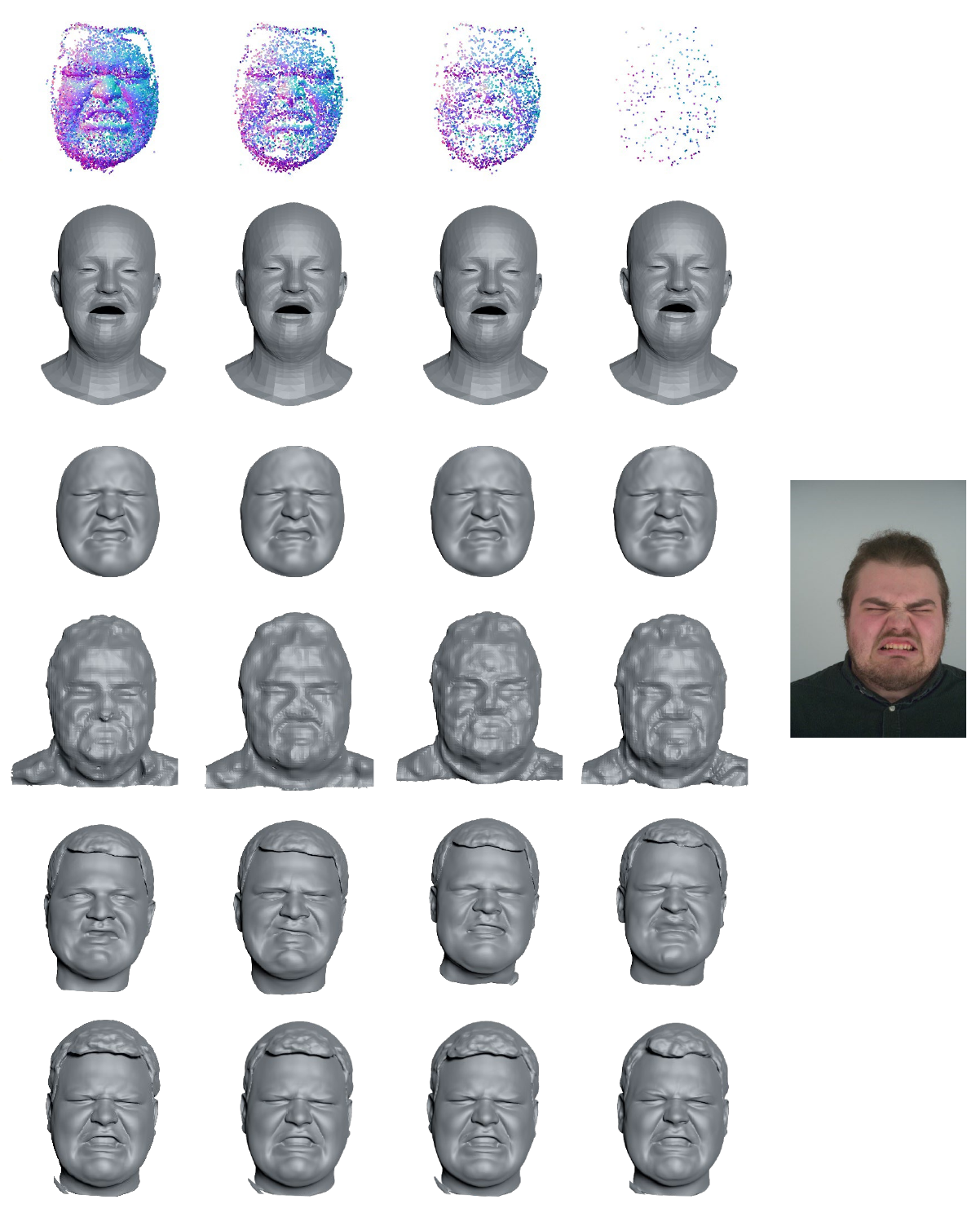}}\\
                \multicolumn{1}{c}{Input} & \\
             \multicolumn{1}{c}{Scans} & \\
              \multicolumn{1}{c}{} & \\
                \multicolumn{1}{c}{} & \\
                \multicolumn{1}{c}{FLAME} & \\
              \multicolumn{1}{c}{} & \\
                \multicolumn{1}{c}{} & \\
                \multicolumn{1}{c}{ImFace*} & \\
              \multicolumn{1}{c}{} & \\
                \multicolumn{1}{c}{} & \\
                \multicolumn{1}{c}{ImAvatar} & \\
              \multicolumn{1}{c}{} & \\
                \multicolumn{1}{c}{} & \\
                \multicolumn{1}{c}{NPHM} & \\
              \multicolumn{1}{c}{} & \\
              \multicolumn{1}{c}{} & \\
                \multicolumn{1}{c}{Ours} & \\
              \multicolumn{1}{c}{} & \\
              \multicolumn{1}{c}{} & \\
    \end{tabular}
    \begin{tabular}{p{35pt}p{28pt}p{28pt}p{25pt}p{28pt}p{28pt}}
                 \#Points
                & 10,000
                & 5,000
                & 2,500
                & 250
                &  RGB
    \end{tabular}
    \caption{Qualitative comparisons of NPHM and our method
with respect to the number of points in the input scan.}
    \label{fig:robust_sparse_supple}
\end{figure}

In Figure~\ref{fig:robust_noise_supple} and~\ref{fig:robust_sparse_supple}, we provide qualitative comparisons against NPHMs on imperfect observations with different noise and sparsity levels.
Detailed quantitative results from our robustness analysis are summarized in Table~\ref{tab:robust_noise} and Table~\ref{tab:robust_sparsity}. 
The results illustrate the superior robustness of DPHMs compared to NPHMs across a range of imperfect observations. 
%
%
\begin{table}[!htbp]
	\renewcommand\arraystretch{1.2}
        \setlength{\tabcolsep}{2.3pt}
	\begin{center}
        \footnotesize
		\begin{tabular}{*{7}{c}}
                \toprule
            \multirow{1}*{Method} & FLAME & ImFace & ImFace*   & ImAvatar & NPHM & Ours 	\\
			\midrule
                0 mm   & 21.01  & 40.71 & 43.71 & 14.84 & 84.49 &  \bf 89.87 \\
                0.5mm  & 21.06  & 40.61 & 43.65 & 14.01  & 82.93 & \bf  88.41  \\
                1mm   & 21.05  & 40.57  & 43.42 & 14.35  & 79.67 & \bf  86.12 \\
                2mm    & 21.08  & 39.86 & 42.64 & 13.85  & 69.29 & \bf  79.41  \\
                \bottomrule
        \end{tabular}
        \caption{Quantitative results at different noise levels of
the input point cloud at each frame.} 
        \label{tab:robust_noise}
        \end{center}
        \vspace{-4mm}
\end{table}

\begin{table}[!htbp]
        \vspace{-4mm}
	\renewcommand\arraystretch{1.2}
        \setlength{\tabcolsep}{2pt}
	\begin{center}
        \footnotesize
		\begin{tabular}{*{10}{c}}
                \toprule
			\multirow{1}*{Method} & FLAME & ImFace & ImFace*   & ImAvatar & NPHM & Ours 	\\
			\midrule
                10,000   & 21.01   & 40.74  & 43.69 & 14.96   & 84.25 & \bf 89.62  \\
                5,000   & 21.06  & 40.71  & 43.72  & 14.84  & 84.49 & \bf 89.87 \\
                2,500   & 21.05   & 40.68  & 43.70  & 14.53    & 83.55 & \bf 89.65 \\
                250   & 21.08  & 40.30  & 42.86 & 14.37   & 81.78 & \bf 85.31 \\
                \bottomrule
        \end{tabular}
        \caption{Quantitative results at different sparsity levels of
the input point cloud at each frame.} 
        \label{tab:robust_sparsity}
        \end{center}
        \vspace{-4mm}
\end{table}

%

\section{Discussions}
\label{SecDiscuss}

%
\paragraph{Hair reconstruction.}
When hair is sufficiently observed, reconstructed hair can be aligned with input (Fig. 1 and second example of Fig. 4 in the main paper). When given a few hair measurements in the depth scans (first example of Fig. 4 in the main paper), we can still output plausible results compared to baselines. We believe that incorporating RGB images with depth scans as inputs could further improve the results.

\paragraph{Arbitrary Length of Sequences.}
Our method formulates head tracking as an optimization problem. It directly optimizes the identity and expression parametric latent. This eliminates the need for an encoder to process input depth scans. 
As mentioned in Sec.~\ref{SecImple}, our non-rigid tracking consists of three stages: identity fitting using the first frame, frame-by-frame expression fitting, and joint fine-tuning of rigid and non-rigid registration parameters.
In the final stage, if the sequence length is short, we can jointly fine-tune the parameters of all frames. However, for long sequences, we can improve temporal smoothness by randomly sampling fragments, with each fragment comprising three consecutive frames. This strategy effectively mitigates memory consumption issues.
Therefore, our method is capable of handling depth sequences of arbitrary length.

\begin{figure}[!htp]
    \centering
    \includegraphics[width=.88\linewidth]{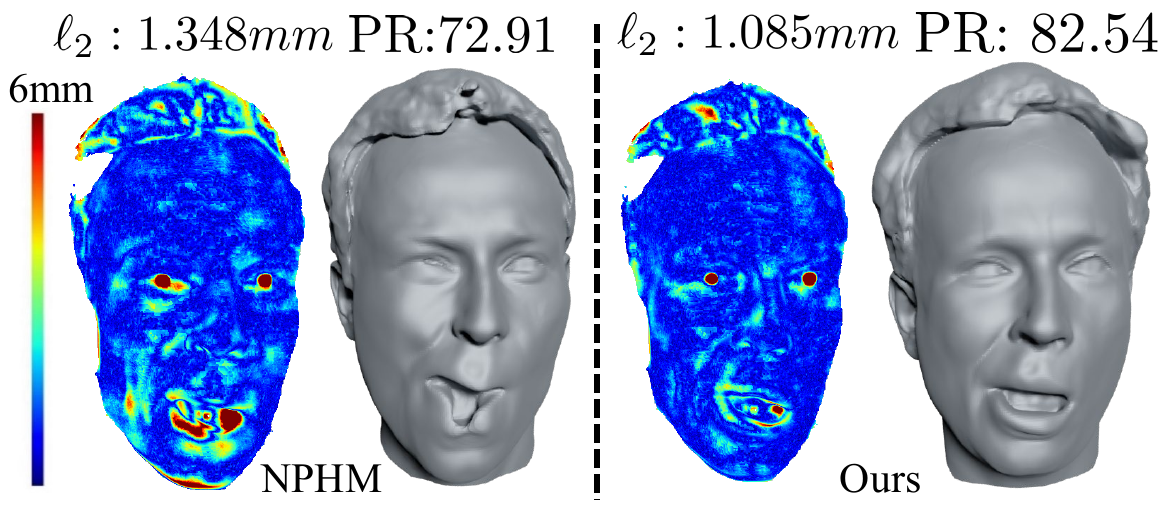}
    \caption{Comparisons between NPHM and our method in reconstructed meshes and error maps from Scan2Mesh distances, along with the calculated metrics. }
    \label{fig:compare_error}
\end{figure}

\paragraph{Non-marginal quantitative improvements.}
In Fig.~\ref{fig:compare_error}, We provide comparisons between NPHM and our method in terms of reconstructed meshes and error maps derived from Scan2Mesh distances, along with the calculated metrics. Notably, 0.263 mm lower $\ell_2$ error means significant improvements in reconstructing facial wrinkles and mouth regions, also with 9.63\% higher Precision@1.5mm score.

\begin{figure}[h] 
    \centering
    \includegraphics[width=\linewidth]{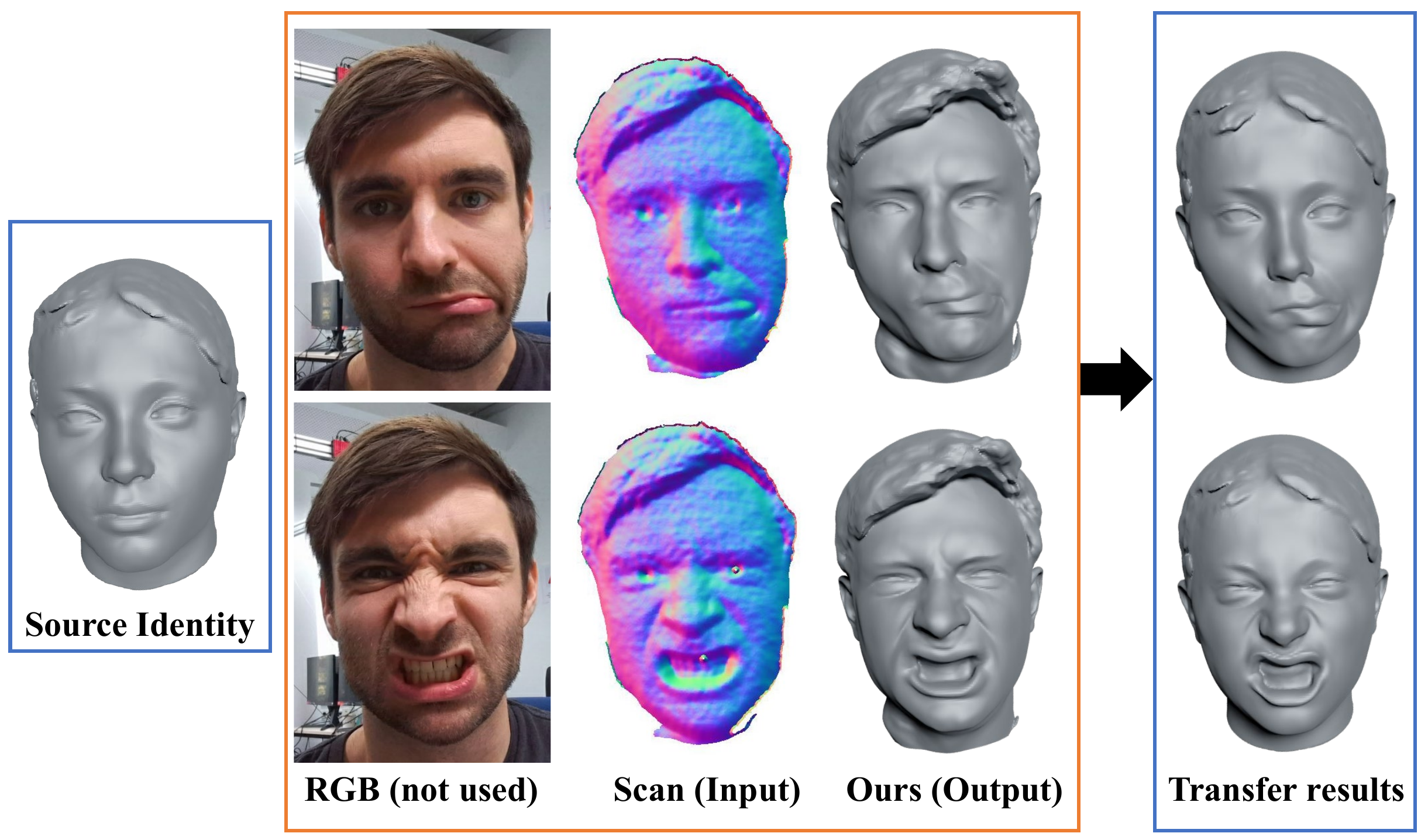}
    \caption{\textbf{Expression Transfer}. Given a monocular depth sequence of a head avatar, we first use our tracking method to obtain the identity and expression latent. Then we transfer the reconstructed expression to the source identity using backward deformation fields, which are conditioned on both the source identity latent and the target expression latent.}
    \label{fig:expre_transfer}
\end{figure}

\paragraph{Expression Transfer.}
As our approach disentangles identity and expression via two separate latents, it can be applied to expression transfer applications.
In Fig.~\ref{fig:expre_transfer}, we demonstrate the transfer of our reconstructed expressions to a different person.  Given a monocular sequence of depth scans as inputs, we initially obtain the identity reconstruction and track expression transitions by our method. Subsequently, we animate the source identity using the reconstructed expression latent.  The transfer result faithfully represents the intricate facial expressions without introducing personalized geometry details such as hairstyle.

\end{document}